\documentclass[10pt,twocolumn,letterpaper]{article}

\usepackage{cvpr}              
\definecolor{cvprblue}{rgb}{0.21,0.49,0.74}
\usepackage[pagebackref,breaklinks,colorlinks,allcolors=cvprblue]{hyperref}

\def\paperID{} 
\def\confName{CVPR}
\def\confYear{2026}

\usepackage{multirow}
\usepackage[normalem]{ulem}
\usepackage{algorithm}
\usepackage{algorithmic}

\usepackage{amsmath,amsfonts}

\usepackage{tabularx}
\usepackage{booktabs}
\usepackage{multirow}
\usepackage[table,xcdraw]{xcolor}
\usepackage{caption}
\usepackage{graphicx}
\usepackage{amsmath}

\title{Towards Highly Transferable Vision-Language Attack via Semantic-Augmented Dynamic Contrastive Interaction}

\author{
Yuanbo Li\textsuperscript{1},
Tianyang Xu\textsuperscript{1},
Cong Hu\textsuperscript{1},
Tao Zhou\textsuperscript{1},
Xiao-Jun Wu\textsuperscript{1}\thanks{Corresponding Author}\,,
Josef Kittler\textsuperscript{2}\\
\small\textsuperscript{1}School of Artificial Intelligence and Computer Science, Jiangnan University\\
\small\textsuperscript{2}Centre for Vision, Speech and Signal Processing (CVSSP), University of Surrey\\
{\tt\small liyuanbo12138@163.com, \{tianyang.xu, conghu, taozhou.ai, wu\_xiaojun\}@jiangnan.edu.cn} \\
{\tt\small j.kittler@surrey.ac.uk}
}

\begin{document}
\maketitle

\begin{abstract}
With the rapid advancement and widespread application of vision-language pre-training (VLP) models, their vulnerability to adversarial attacks has become a critical concern. 
In general, the adversarial examples can typically be designed to exhibit transferable power, attacking not only different models but also across diverse tasks.
However, existing attacks on language-vision models mainly rely on static cross-modal interactions and focus solely on disrupting positive image-text pairs, resulting in limited cross-modal disruption and poor transferability.
To address this issue, we propose a Semantic-Augmented Dynamic Contrastive Attack (SADCA) that enhances adversarial transferability through progressive and semantically guided perturbation.
SADCA progressively disrupts cross-modal alignment through dynamic interactions between adversarial images and texts.
This is accomplished by SADCA establishing a contrastive learning mechanism involving adversarial, positive and negative samples, to reinforce the semantic inconsistency of the obtained perturbations.
Moreover, we empirically find that input transformations commonly used in traditional transfer-based attacks also benefit VLPs, which motivates a semantic augmentation module that increases the diversity and generalization of adversarial examples.
Extensive experiments on multiple datasets and models demonstrate that SADCA significantly improves adversarial transferability and consistently surpasses state-of-the-art methods.
The code is released at \url{https://github.com/LiYuanBoJNU/SADCA}.
\end{abstract}

\begin{figure}[t!]
\centering
\includegraphics[width=0.99\columnwidth]{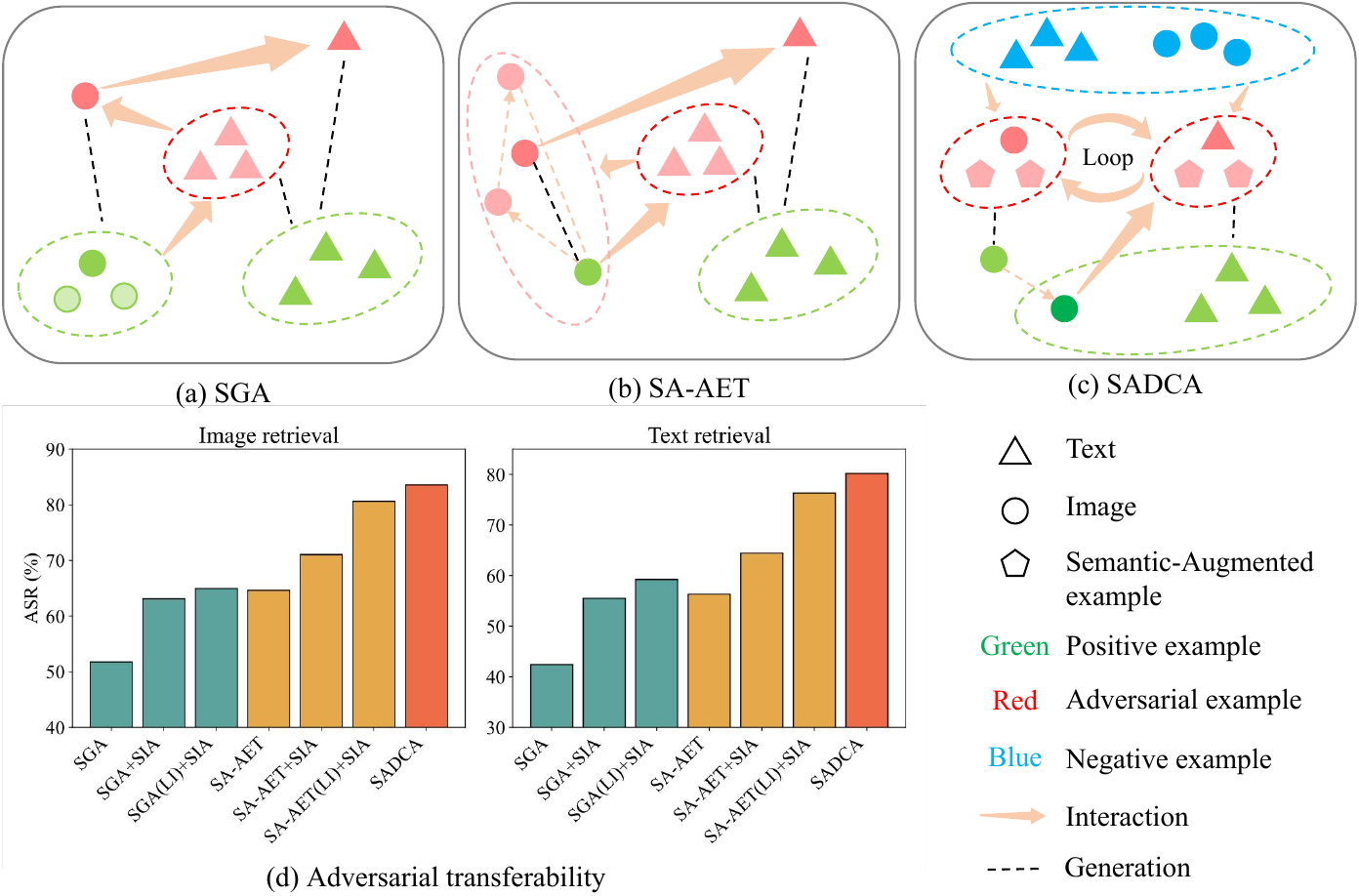}
\caption{A comparison of our SADCA and existing frameworks. (a) and (b) illustrate the core concepts of SGA~\cite{lu2023set} and SA-AET~\cite{jia2024semantic}, respectively, where only one or two static interactions are performed between the visual and textual modalities, with the interactions being limited solely to positive pairs. (c) illustrates the core idea of the proposed SADCA, which continuously disrupts cross-modal interactions through dynamic contrastive interactions with both positive and negative pairs. Additionally, it leverages a semantic augmentation strategy to enrich the data samples, thereby diversifying the semantic information. The arrow represents the interaction between the visual and textual modalities. The dotted lines represent the generation of adversarial examples from the original examples. (d) demonstrates the effectiveness of the input transformation~\cite{wang2023structure} in enhancing the adversarial attack transferability. Furthermore, we observe that using large number of iterations (LI) to attack the image modality can further improve the attack performance.}
\label{Motivation}
\end{figure}

\section{Introduction}
Vision-language pre-training (VLP) models learn cross-modal semantic correlations through joint training on large-scale image-text pairs, providing superior feature representations and significantly enhancing performance across various tasks, such as image-text retrieval (ITR)~\cite{huang2024cross, kang2025calibclip}, image captioning (IC)~\cite{zeng2024meacap, hua2025finecaption}, and visual grounding (VG)~\cite{wang2025enhancing,zhang2024vision}. 
However, their robustness and reliability are of increasing concern, as numerous studies have demonstrated that VLP models are highly susceptible to adversarial attacks~\cite{liu2024divide, zhang2022towards,qi2024visual,zhao2023evaluating,xu2024highly, kong2024patch}. 
Investigating the generation of adversarial examples to attack VLP models is instrumental to the design of more robust VLP architectures, thus ensuring a secure deployment of foundational models.

Previous studies have primarily focused on generating adversarial examples for VLP models in white-box settings~\cite{zhang2022towards}, where the attacker has access to internal model information such as parameters. 
However, in real-world scenarios, obtaining such detailed information is often infeasible, which limits the practicality of white-box attacks. 
This limitation has motivated increasing attention toward transfer-based attacks, which craft adversarial examples using known surrogate models and then transfer them to attack unseen black-box targets~\cite{wang2025exploring}.
Several approaches have achieved certain success in improving the adversarial transferability to VLP models, such as SGA~\cite{lu2023set}, DRA~\cite{gao2024boosting}, and SA-AET~\cite{jia2024semantic}. 
Nevertheless, these approaches rely on static cross-modal interactions, generating adversarial examples solely from the original image-text pair, where the interaction between modalities occurs only once or twice.
Such a design causes the generated adversarial examples to deviate from the semantic center along a fixed direction, lacking the ability to explore diverse directions within the semantic space. 
As a result, these methods struggle to disrupt the visual-textual alignment, which limits their generalization, as well as the cross-model transferability of the generated adversarial examples.
Moreover, existing attacks typically involve only positive image-text pairs, neglecting the role of negative samples in shaping the semantic decision boundary.
This one-sided optimization produces only a repulsive force that pushes examples away from their original semantic cluster but does not provide the complementary attractive force needed to pull them across semantic boundaries.
Consequently, the adversarial examples remain inadequately separated from benign samples in the embedding space, thereby weakening their transferability.
The reliance on static cross-modal interactions, coupled with the neglect of negative samples, significantly limits the transferability of existing vision-language adversarial attacks, as shown in Figure \ref{Motivation}(a), (b), and (d).

Additionally, much like the efforts to enhance the generalization ability of deep neural networks (DNNs), increasing data diversity plays a crucial role in improving adversarial transferability~\cite{zhu2022toward}.
Input transformations, as an effective strategy for enhancing data diversity, have been extensively studied and proven to improve the transferability of adversarial examples in traditional attack settings~\cite{wang2023structure,wang2024boosting} significantly.
However, existing vision-language attacks largely overlook the importance of input transformations, often considering only limited scale-invariance within the image modality~\cite{lu2023set}. 
This lack of comprehensive data augmentation results in insufficient input diversity, thereby hindering the ability to achieve strong adversarial transferability.

In this paper, to achieve highly transferable vision-language attacks, we propose a novel Semantic-Augmented Dynamic Contrastive Attack (SADCA) to improve the transferability of adversarial examples.
Specifically, as illustrated in Figure \ref{Motivation}(c), SADCA introduces a dynamic contrastive interaction mechanism that iteratively disrupts cross-modal semantic consistency by alternately updating adversarial images and texts.
Simultaneously, it incorporates a contrastive learning strategy that maximizes similarity with negative samples while minimizing the alignment with positive samples. 
This effectively breaks the semantic matching relationships between images and texts, enhancing the  capability of adversarial examples to be semantically misleading and improving their transferability.
In addition, as shown in Figure \ref{Motivation}(d), we empirically investigate the effectiveness of integrating existing input transformation strategies into vision-language attacks, and, in addition, propose a semantic augmentation module. 
This module performs local semantic enhancement on images and mixed semantic augmentation on texts to broaden the set of semantic views encountered during optimization.
The resulting diversity yields richer semantic gradients and reduces overfitting to a single view, which together improve the generalization and cross-model transferability of the generated adversarial examples.
Extensive experiments on multiple datasets and VLP models show that SADCA consistently outperforms state-of-the-art (SOTA) methods in adversarial transferability.
Our contributions can be summarized as follows:
\begin{itemize}
    \item We propose SADCA, a novel highly transferable vision-language attack method that iteratively disrupt image–text semantic consistency and amplify cross-modal misalignment by dynamic contrastive interaction.
    \item SADCA further introduces a semantic augmentation module to diversify the semantic information of adversarial examples, obtaining richer semantic gradients to further enhance adversarial transferability.
    \item Experimental results show that SADCA consistently outperforms SOTA approaches in cross-model and cross-task adversarial transferability.
\end{itemize}

\section{Related Work}
\subsection{Vision-Language Pre-training Models}
Vision-Language Pretraining (VLP) models are a class of multimodal learning frameworks designed to jointly model visual and textual modalities~\cite{tschannen2025siglip, sun2023eva, li2022blip, wang2024cogvlm, bai2025qwen2}. Their primary objective is to perform pretraining on large-scale image-text pair datasets to learn generalizable cross-modal representations, which can be effectively transferred to a variety of downstream tasks.
Based on their architectural design, VLP models can be broadly classified into fusion-based and alignment-based models. Fusion-based models (\textit{e.g.}, ALBEF~\cite{li2021align}, TCL~\cite{yang2022vision}) typically use separate unimodal encoders to independently extract textual token embeddings and visual features. These are subsequently fused by a multimodal encoder to generate unified semantic representations. In contrast, alignment-based models (\textit{e.g.}, CLIP~\cite{radford2021learning}) adopt two independent unimodal encoders for images and texts, aligning their embeddings within a shared semantic space through contrastive learning.

\subsection{Transferable Adversarial Attacks}
Transferable adversarial attacks~\cite{ren2025improving, li2025transferable, xie2025chain, li2025hierarchical} aim to generate adversarial examples on a white-box (source) model that can successfully mislead an unseen black-box (target) model. Due to their black-box nature, such attacks are more applicable in real-world scenarios and are considered more valuable for research than white-box attacks.
Input transformation is a widely adopted technique in image recognition to enhance the transferability of adversarial examples, with many methods having been proposed, such as Admix~\cite{wang2021admix}, SIA~\cite{wang2023structure}, and BSR~\cite{wang2024boosting}. 
Recently, studies have begun exploring transferable adversarial attacks in VLP models~\cite{wang2024transferable,he2023sa,han2023ot,zhang2025maa,wang2025exploring}.
SGA~\cite{lu2023set} improves the transferability of multimodal adversarial examples by expanding a single image-text pair into a diverse set, thereby increasing adversarial diversity. DRA~\cite{gao2024boosting} exploits the intersection area of adversarial trajectories during optimization to enrich adversarial diversity and reduce the risk of overfitting. Furthermore, SA-AET~\cite{jia2024semantic} refines adversarial trajectories and generates adversarial examples in a contrastive space of semantic image-text features.
However, these methods are constrained by static cross-modal interactions and the absence of guidance from negative pairs, making it difficult to fully disrupt the cross-modal alignment of adversarial examples. Additionally, their limited use of input transformations increases the risk of model overfitting and reduces the transferability of adversarial examples.

\section{Methodology}
\subsection{Notations and Preliminaries}
Let $(v, t)$ denote the benign image-text pair extracted from a multimodal dataset $D$. For VLP model, the $F_I(\cdot)$ and $F_T(\cdot)$ represent the image encoder and text encoder, respectively.
Vision-language attacks typically induce a semantic mismatch between the adversarial image and its corresponding adversarial text, while adhering to predefined perturbation constraints in both the visual and textual domains. Given a paired image and text input, the objective is to introduce subtle adversarial perturbations to both modalities such that the semantic similarity between the image and its associated text is significantly reduced. To achieve this, multimodal adversarial examples are generated by minimizing a feature similarity loss function:
\begin{equation}
\label{Preliminaries}
    \left\{
    \begin{array}
    {l}\min J\left(F_I\left(v'\right),F_T\left(t'\right)\right) \\
    \mathrm{s.t.} \ v'\in B\left[v,\epsilon_v\right],t'\in B\left[t,\epsilon_t\right]
    \end{array}\right.
\end{equation}
where $v'$ and $t'$ denote the adversarial image-text pair within the image and text search spaces $B\left[v,\epsilon_v\right]$ and $B\left[t,\epsilon_t\right]$, $\epsilon_v$ and $\epsilon_t$ represent the maximum perturbation budgets for the image and text modalities, respectively. In this paper, we adopt cosine similarity $Cos(v',t')=\frac{F_I(v')\cdot F_T(t')}{||F_I(v')||\cdot||F_T(t')||}$ to measure the feature similarity of image-text pairs.
For the adversarial image $v'$, the added perturbation should preserve visual similarity to the benign image $v$ under the given constraints. For the adversarial text $t'$, modifications are made to the words in the benign text $t$ without altering the original semantic meaning as perceived by humans.

\subsection{SADCA}
Despite recent progress in vision-language adversarial attacks, existing methods exhibit fundamental limitations that hinder transferability. They rely on static cross-modal interactions, generating adversarial examples in fixed semantic directions, and focus mainly on positive samples, neglecting negative samples that shape semantic boundaries. Consequently, adversarial examples experience limited separation from benign examples. Moreover, input diversity and data augmentation are underutilized, restricting coverage of the semantic space. These limitations motivate a more dynamic, contrastive, and semantically aware attack framework.
To address this, we propose SADCA that aligns a benign image with multiple textual descriptions to obtain a semantically centered representation, then uses both positive and negative pairs in an iterative contrastive framework to induce continuous semantic misalignment. By pushing samples away from the semantic center and pulling them across negative samples, SADCA generates adversarial examples that are semantically more divergent and highly transferable across diverse VLP models and tasks.

\textbf{Dynamic Contrastive Interaction.}
Since image feature embeddings often contain a substantial amount of information that is irrelevant to the corresponding text features, this redundant information can distort similarity computations and cause misalignment between visual and textual representations~\cite{jia2024semantic}. 
Using such biased original image-text pairs as positive samples directly may prevent adversarial examples from being displaced from the correct semantic center, thereby limiting their transferability. 
To address this, we first align the benign image and text representations within the semantic space to obtain a semantically centered aligned image, which is then used as the positive sample for subsequent image-text interactions.
Specifically, we maximize the feature similarity between the benign image and the benign text to obtain a semantically centered positive image:
\begin{equation}
\label{aligned}
v_p=\underset{v_p \in B\left[v, \epsilon_{v}\right]}{\arg \max }\sum_{m=1}^{M} Cos(v,t_m),
\end{equation}
where $t_m\in T=\{t_1,t_2...,t_M\}$ is the text caption paired with the benign image, and $T$ represents the set of multiple text captions associated with the benign image. 
By aligning with multiple text features in the semantic space, a positive image $v_p$ representation, that is closer to the semantic center, can be obtained.
For the positive text set, the benign text set $T$ is directly used as the positive text set $T_p=\{t_{p1},t_{p2}...,t_{pM}\}$.

For the selection of negative samples, we randomly choose $K$ mismatched samples from the dataset to construct the negative image set $V_n=\{v_{n1},v_{n2}...,v_{nK}\}$ and the negative text set $T_n=\{t_{n1},t_{n2}...,t_{nK}\}$.

Leveraging the constructed positive and negative pairs, our objective is to guide the semantic deviation of adversarial examples away from the benign semantic center through a contrastive learning framework. By minimizing the similarity between adversarial examples and positive samples, while maximizing their similarity with negative samples in the semantic embedding space, we induce semantic-level shifts and ambiguity in the adversarial perturbations. This strategy enhances cross-modal adversarial transferability.
Specifically, we minimize the feature similarity between adversarial samples and positive pairs while maximizing the feature similarity with negative pairs:
\begin{equation}
\small
\label{contrastive_v}
\min_{v_i'} \mathcal{L}\left(v_i', T_p, T_n\right)= \sum_{m=1}^{M} Cos(v_i',t_{pm}) - \lambda  \sum_{k=1}^{K} Cos(v_i',t_{nk})
\end{equation}
\begin{equation}
\small
\label{contrastive_t}
\min_{t_i'}\mathcal{L}\left(t_i', v_p, V_n\right)= Cos(v_p,t_i') - \lambda  \sum_{k=1}^{K} Cos(v_{nk},t_i')
\end{equation}
where $v'_i$ and $t'_i$ denote the adversarial image and adversarial text at the $i$-th iteration, $\lambda$ is the weighting factor that controls the influence of negative samples in the guidance process, $\mathcal{L}\left(v_i', T_p, T_n\right)$ and $\mathcal{L}\left(t_i', v_p, V_n\right)$ denote the loss functions to optimize the adversarial image $v_i'$ and the adversarial text $t_i'$, respectively.

Moreover, to thoroughly disrupt the cross-modal interactions of adversarial examples, we adopt a dynamic interaction mechanism that continuously perceives and disrupts the semantic alignment between adversarial images and texts during the attack process.
Specifically, we minimize the feature similarity between the current adversarial image and adversarial text throughout the iterative process, thereby progressively weakening their semantic consistency:
\begin{equation}
\small
\label{contrastive_dv}
\min_{v'_i} \mathcal{L}\left(v'_i, T'_i, T_n\right)= \sum_{m=1}^{M} Cos(v'_i,t'_{im}) - \lambda  \sum_{k=1}^{K} Cos(v'_i, t_{nk})
\end{equation}
\begin{equation}
\small
\label{contrastive_dt}
\min_{t'_i}\mathcal{L}\left(t'_i, v'_i, V_n\right)= Cos(v'_i,t'_i) - \lambda  \sum_{k=1}^{K} Cos(v_{nk}, t'_i)
\end{equation}
where $v'_i$ and $t'_i$ denote the adversarial image and adversarial text at the $i$-th iteration. 

The dynamic contrastive interaction operates on the current adversarial image and text at each iteration, meaning that the semantic representations of both modalities are continuously updated and the resulting gradient directions change accordingly. In each round of interaction, the degree of repulsion from positive samples and attraction toward negative samples is recalibrated, based on the latest semantic states, causing the adversarial examples to exhibit continuous semantic drift in the semantic space and to explore a broader range of potential attack directions. This iterative, dynamic adjustment not only enhances gradient diversity but also substantially improves transferability across models and tasks.

\begin{algorithm}[t]
\caption{SADCA}
\label{SADCA}
\textbf{Input}: Image encoder $F_I$, Text encoder $F_T$, Dataset $D$, Image-text pair $\{v,t\}$, number of paired text $M$, iteration steps of adversarial image generation $J$, step size $\alpha$, momentum factor $\mu$, dynamic interaction steps $I$, number of negative samples $K$, weighting factor of negative samples $\lambda$, semantic augmentation number $S$.\\
\textbf{Output}: Adversarial image $v'$ and adversarial text $t'$.

\begin{algorithmic}[1] 
\STATE Build text set $T=\{t_1,t_2...,t_M\}\gets D$
\STATE Build negative image set $V_n$ and negative text set $T_n$
\STATE $v_p={\arg \max }\sum_{m=1}^{M} Cos(v,t_m)$.
\STATE $v'_0=v,T'_0=T,g_0=0$
\FOR{$i=0$ to $I-1$}
\STATE $\mathcal{L}_t=\mathcal{L}(t'_{im},v'_i,I_n)+\mathcal{L}(t'_{im},v_p,I_n)$
\STATE $T'_{i+1}=\{t'_{(i+1)m}={\arg \min } \mathcal{L}_t\}^M_{m=1}$
\FOR{$j=0$ to $J-1$}
\STATE Generate $V'_{sa}$ from $v'_{ij}$ by Eq.~(\ref{Image SA})
\STATE Generate $T'_{sa}$ from $T'_{i+1}$ by Eq.~(\ref{Text SA})
\STATE $\mathcal{L}_v=\mathcal{L}\left(V'_{sa}, T_p, T_n\right)+\mathcal{L}\left(V'_{sa}, T'_{sa}, T_n\right)$
\STATE $g_{i(j+1)}=\mu \cdot g_{ij} + \frac{\nabla_{\mathcal{L}_v}}{\left\|\nabla_{\mathcal{L}_v}\right\|}$
\STATE $v'_{i(j+1)}=clip_{v,\epsilon_v}(v'_{ij}+\alpha \cdot \text{sign}(g_{i(j+1)})$
\ENDFOR
\ENDFOR
\STATE \textbf{return} $v'=v'_{IJ}$, $t'=t'_{I}$
\end{algorithmic}
\end{algorithm}

\textbf{Semantic Augmentation Module.}
Although input transformation strategies originally developed for traditional image-based adversarial attacks have proven effective in improving the generalization of vision-language attacks, they primarily operate on the perceptual-level image features and overlook the cross-modal semantic alignment mechanisms that are unique to VLP models. As a result, these transformations struggle to disrupt the semantic matching process between image and text, leading to limited cross-modal transferability and weakened cross-modal attack effectiveness.
To address this limitation, we propose the semantic augmentation module, which aims to enhance the diversity of semantic structures in the input, while preserving the semantic consistency of the original image-text labels. This approach enables more effective disruption of the alignment relationships within VLP models.

The semantic augmentation module consists of local semantic image augmentation and mixed semantic text augmentation. Specifically, local semantic image augmentation amplifies local semantic regions within the image, guiding the attack to focus on finer-grained semantic information. Given the adversarial image $v'$, we construct a set of locally semantic-augmented images $V'_{sa}$, where each image is generated by randomly cropping and resizing local regions. Additionally, a random image augmentation function $A\in \mathcal{A}=\{Rotate,Brightness,Flip...\}$ is applied to each image, selected from a set of diverse augmentation operations.
The local semantic image augmentation process can be formalized as:
\begin{equation}
\label{Image SA}
V'_{sa}=\{A_{s}\left(Resize\left(Crop\left(v';r_{s}\right)\right)\right)\}^S_{s=1},
\end{equation}
where $r_{s} \sim \mathcal{U}(0.4,0.8)$ denotes the ratio factor, representing the area of the cropped region relative to the area of the original image. The mixed semantic text augmentation enhances semantic diversity by combining multiple textual descriptions to form broader semantic representations, thereby promoting inconsistency between image and text features in the semantic space. Specifically, it randomly selects and concatenates pairs of text samples from the adversarial text set $T'$, resulting in a new augmented text set $T'_{sa}$:
\begin{equation}
\label{Text SA}
T'_{sa}=\{t_s=Concat(t'_i,t'_j)|t'_i,t'_j\in T',i\neq j\}^S_{s=1}.
\end{equation}
The semantic augmentation module increases the diversity of semantic representation, while preserving overall semantic consistency, enabling the attack to disrupt cross-modal interactions within a broader semantic space. This helps to mitigate overfitting and enhances the transferability of adversarial examples.
The overall procedure of the proposed SADCA is summarized in Algorithm \ref{SADCA}.

\begin{table*}[t!]
\caption{A comparison of SADCA with SOTA methods on the image-text retrieval (ITR) task using the Flickr30K dataset. The "Source" column indicates the VLP model used to generate the multimodal adversarial examples. For both image retrieval (IR) and text retrieval (TR), we report the ASR (\%) at Rank-1 (R@1). The "Average" represents the average ASR on the black-box VLP models.}
\centering
\small
\setlength{\tabcolsep}{3pt}
\begin{tabular}{c|c|cc|cc|cc|cc|cc}
\midrule
\multirow{2}{*}{\textbf{Source}} & \multirow{2}{*}{\textbf{Attack}} & \multicolumn{2}{c|}{\textbf{ALBEF}} & \multicolumn{2}{c|}{\textbf{TCL}} & \multicolumn{2}{c|}{\textbf{CLIP\textsubscript{ViT}}} & \multicolumn{2}{c|}{\textbf{CLIP\textsubscript{CNN}}}  & \multicolumn{2}{c}{\textbf{Average}} \\
 &  & TR R@1 & IR R@1 & TR R@1 & IR R@1 & TR R@1 & IR R@1 & TR R@1 & IR R@1 & TR R@1 & IR R@1 \\
\midrule
\multirow{9}{*}{\textbf{ALBEF}} 
& PGD & 93.74 & 94.43 & 24.03 & 27.90 & 10.67 & 15.82 & 14.05 & 19.11 & 16.25 & 20.94 \\
& BERT-Attack & 11.57 & 27.46 & 12.64 & 28.07 & 29.33 & 43.17 & 32.69 & 46.11 & 24.89 & 39.12 \\
& Co-Attack & 97.08 & 98.36 & 39.52 & 51.24 & 29.82 & 38.92 & 31.29 & 41.99 & 33.54 & 44.72 \\
& SGA & 99.90 & 99.93 & 87.88 & 88.05 & 36.69 & 46.78 & 39.59 & 49.78 & 54.72 & 61.54 \\
& SGA(LI)+SIA & 100 & 100 & 99.37 & 99.02 & 49.82 & 56.19 & 54.79 & 60.99 & 67.99 & 72.07 \\
& DRA & 99.90 & 99.93 & 91.57 & 91.36 & 46.26 & 56.80 & 49.55 & 59.01 & 62.46 & 69.06 \\
& SA-AET & 99.90 & 99.98 & 96.42 & 96.02 & 55.58 & 63.89 & 57.22 & 65.59 & 69.74 & 75.17 \\
& SA-AET(LI)+SIA & 100 & 100 & \textbf{99.58} & \textbf{99.38} & 75.71 & 78.58 & 76.25 & 80.41 & 83.85 & 86.12\\
& \cellcolor{gray!15}SADCA (ours) & \cellcolor{gray!15}\textbf{100} & \cellcolor{gray!15}\textbf{100} & \cellcolor{gray!15}98.52 & \cellcolor{gray!15}97.83 & \cellcolor{gray!15}\textbf{81.10} & \cellcolor{gray!15}\textbf{82.83} & \cellcolor{gray!15}\textbf{85.44} & \cellcolor{gray!15}\textbf{86.11} &
\cellcolor{gray!15}\textbf{88.35} & \cellcolor{gray!15}\textbf{88.92} \\
\midrule
\multirow{9}{*}{\textbf{TCL}} 
& PGD & 35.77 & 41.67 & 99.37 & 99.33 & 10.18 & 16.30 & 14.81 & 21.10 & 20.25 & 26.36 \\
& BERT-Attack & 11.89 & 26.82 & 14.54 & 29.17 & 29.69 & 43.36 & 33.46 & 46.07 & 25.01 & 38.75 \\
& Co-Attack & 49.84 & 60.36 & 91.68 & 93.65 & 32.64 & 42.69 & 32.06 & 47.82 & 38.18 & 50.29 \\
& SGA & 92.40 & 92.77 & 100 & 100 & 36.81 & 46.97 & 41.89 & 51.53 & 57.70 & 63.76 \\
& SGA(LI)+SIA & 99.90 & 99.70 & 100 & 100 & 52.87 & 59.12 & 62.45 & 65.49 & 71.74 & 74.77 \\
& DRA & 95.20 & 95.58 & 100 & 100 & 47.24 & 57.28 & 52.23 & 62.23 & 64.89 & 71.70 \\
& SA-AET & 98.85 & 98.50 & 100 & 100 & 56.20 & 63.47 & 59.77 & 67.86 & 71.61 & 76.61 \\
& SA-AET(LI)+SIA & \textbf{99.95} & \textbf{99.93} & 100 & 100 & 77.04 & 81.48 & 80.20 & 84.05 & 85.73 & 88.49\\
& \cellcolor{gray!15}SADCA (ours) & \cellcolor{gray!15}99.58 & \cellcolor{gray!15}99.56 & \cellcolor{gray!15}\textbf{100} & \cellcolor{gray!15}\textbf{100} & \cellcolor{gray!15}\textbf{78.28} & \cellcolor{gray!15}\textbf{83.18} & \cellcolor{gray!15}\textbf{86.46} & \cellcolor{gray!15}\textbf{88.71}
& \cellcolor{gray!15}\textbf{88.11} & \cellcolor{gray!15}\textbf{90.48}\\
\midrule
\multirow{9}{*}{\textbf{CLIP\textsubscript{ViT}}}
& PGD & 3.13 & 6.48 & 4.43 & 8.83 & 69.33 & 84.79 & 13.03 & 17.43 & 6.86 & 10.91 \\
& BERT-Attack & 9.59 & 22.64 & 11.80 & 25.07 & 28.34 & 39.00 & 30.40 & 37.43 & 17.26 & 28.38 \\
& Co-Attack & 8.55 & 20.18 & 10.01 & 21.29 & 78.53 & 87.50 & 29.50 & 38.49 & 16.02 & 26.65 \\
& SGA & 22.42 & 34.59 & 25.08 & 36.45 & 100 & 100 & 53.26 & 61.10 & 33.59 & 44.05 \\
& SGA(LI)+SIA & 54.33 & 60.20 & 57.43 & 62.60 & 100 & 100 & 89.02 & 89.24 & 66.93 & 70.68 \\
& DRA & 27.84 & 42.84 & 27.82 & 44.60 & 100 & 100 & 64.88 & 69.50 & 40.18 & 52.31 \\
& SA-AET & 36.60 & 50.44 & 39.20 & 51.10 & 100 & 100 & 71.01 & 74.10 & 48.94 & 58.55  \\
& SA-AET(LI)+SIA & 79.04 & 82.74 & 79.35 & 82.57 & 100 & 99.97 & 94.76 & 95.23 & 84.38 & 86.85 \\
& \cellcolor{gray!15}SADCA (ours) & \cellcolor{gray!15}\textbf{87.07} & \cellcolor{gray!15}\textbf{89.20} & \cellcolor{gray!15}\textbf{87.04} & \cellcolor{gray!15}\textbf{87.98} & \cellcolor{gray!15}\textbf{100} & \cellcolor{gray!15}\textbf{100} & \cellcolor{gray!15}\textbf{97.90} & \cellcolor{gray!15}\textbf{97.46} & \cellcolor{gray!15}\textbf{90.68} & \cellcolor{gray!15}\textbf{91.55}\\
\midrule
\multirow{9}{*}{\textbf{CLIP\textsubscript{CNN}}} 
& PGD & 2.29 & 6.15 & 4.53 & 8.88 & 5.40 & 12.08 & 89.78 & 93.04 & 4.07 & 9.04 \\
& BERT-Attack & 9.86 & 23.27 & 12.33 & 25.48 & 27.12 & 37.44 & 30.40 & 40.10 & 16.44 & 28.73 \\
& Co-Attack & 10.53 & 23.62 & 12.54 & 26.05 & 27.24 & 40.62 & 95.91 & 96.50 & 16.77 & 30.10 \\
& SGA & 15.64 & 28.06 & 18.02 & 33.07 & 39.92 & 51.45 & 99.87 & 99.90 & 24.53 & 37.53 \\
& SGA(LI)+SIA & 20.96 & 32.72 & 22.55 & 37.21 & 47.24 & 56.70 & 100 & 100 & 30.25 & 42.21 \\
& DRA & 19.50 & 34.59 & 21.60 & 37.88 & 48.47 & 59.12 & 99.87 & 99.90 & 29.86 & 43.86 \\
& SA-AET & 23.98 & 38.28 & 27.29 & 41.81 & 54.11 & 64.21 & 100 & 99.97 & 35.13 & 48.10 \\
& SA-AET(LI)+SIA & 38.69 & 51.80 & 44.89 & 56.33 & 69.57 & 74.68 & 100 & 100 & 51.05 & 60.94 \\
& \cellcolor{gray!15}SADCA (ours) & \cellcolor{gray!15}\textbf{49.43} & \cellcolor{gray!15}\textbf{60.55} & \cellcolor{gray!15}\textbf{55.53} & \cellcolor{gray!15}\textbf{63.19} & \cellcolor{gray!15}\textbf{77.18} & \cellcolor{gray!15}\textbf{79.57} & \cellcolor{gray!15}\textbf{100} & \cellcolor{gray!15}\textbf{100} & \cellcolor{gray!15}\textbf{60.71} & \cellcolor{gray!15}\textbf{67.77} \\
\midrule
\end{tabular}
\label{cross-model}
\end{table*}

\begin{table*}[t!]
\caption{Cross-task transferability. We use ALBEF model to generate multimodal adversarial examples and apply them to attack the Visual Grounding (VG) task on the RefCOCO+ dataset and the Image Captioning (IC) task on the MSCOCO dataset. The "Clean" denotes the performance of each task without any attack, where lower values indicate better effectiveness of the adversarial attack in both tasks.}
\label{cross-task}
\centering
\resizebox{0.85\textwidth}{!}{
\begin{tabular}{c|ccc|ccccc}
\midrule
\multirow{2}{*}{\textbf{Attack}} & \multicolumn{3}{c|}{\textbf{ITR $\rightarrow$ VG}} & \multicolumn{5}{c}{\textbf{ITR $\rightarrow$ IC}} \\
 & Val $\downarrow$ & TestA $\downarrow$ & TestB $\downarrow$ & B@4 $\downarrow$ & METEOR $\downarrow$ & ROUGE-L $\downarrow$ & CIDEr $\downarrow$ & SPICE $\downarrow$ \\
\midrule
Clean & 58.46 & 65.89 & 46.25 & 39.7 & 31.0 & 60.0 & 133.3 & 23.8 \\
SGA & 50.56 & 57.42 & 40.66 & 28.0 & 24.6 & 51.2 & 91.4 & 17.7 \\
DRA & 49.70 & 56.32 & 40.54 & 27.2 & 24.2 & 50.7 & 88.3 & 17.2 \\
SA-AET & 47.44 & 53.27 & 38.58 & 21.0 & 20.5 & 45.2 & 65.7 & 13.6 \\
\rowcolor{gray!15}
\textbf{SADCA (ours)} & \textbf{46.78} & \textbf{52.41} & \textbf{37.10} & \textbf{17.4} & \textbf{17.9} & \textbf{42.2} & \textbf{50.3} & \textbf{10.7} \\
\midrule
\end{tabular}
}
\end{table*}

\begin{table*}[t!]
\caption{Adversarial transferability on LVLMs. We use ALBEF model to generate multimodal adversarial examples and apply them to attack the LVLMs on the Flickr30K dataset.}
\centering
\resizebox{0.99\textwidth}{!}{
\begin{tabular}{c|cccccccc} 
\midrule
Attack                                                  & LLaVA-1.5-7B     & Qwen3-VL-8B        & InternVL3-8B      & Qwen2.5-VL-7B     & GPT-5 & GPT-4o-mini & Gemini-2.0 & Claude-4.5  \\ 
\midrule
Clean                                                   & 3.46          & 14.4          & 4.72          & 11.04  & 23.88  & 15.00  & 6.96      & 21.80            \\
SGA                                                     & 17.60          & 47.00          & 17.60          & 36.00  & 50.65 & 40.96  & 30.12      & 63.32            \\
SGA(LI)+SIA                                             & 27.30          & 62.76          & 27.30          & 45.92  & 52.17 & 46.54  & 27.92      & 64.18            \\
SA-AET                                                  & 14.30          & 60.16          & 25.72          & 48.20  & 60.26 & 50.86  & 38.50      & 72.28            \\
SA-AET(LI)+SIA                                          & 35.20          & 80.14          & 46.68          & 67.50  & 68.08 & 62.48  & 41.56      & 75.12            \\
\rowcolor[rgb]{0.925,0.925,0.925} \textbf{SADCA (ours)} & \textbf{40.34} & \textbf{86.34} & \textbf{69.06} & \textbf{82.82} & \textbf{78.61} & \textbf{79.12} & \textbf{52.06} & \textbf{79.02}            \\
\midrule
\end{tabular}
}
\label{LVLMs}
\end{table*}

\section{Experiments}
\subsection{Experimental Setting}

\textbf{Dataset and VLP models.}
To evaluate the effectiveness of the proposed method, we select two widely used multimodal datasets, Flickr30K~\cite{plummer2015flickr30k} and MSCOCO~\cite{lin2014microsoft}, to conduct experiments on the image-text retrieval task. In the cross-task transferability experiments, we use the RefCOCO+~\cite{yu2016modeling} dataset for the visual grounding task and the MSCOCO dataset for the image captioning task. Following \cite{lu2023set}, we employ two popular types of VLP models: fusion-based VLP and alignment-based VLP. For fusion-based VLP models, we consider ALBEF~\cite{li2021align} and TCL~\cite{yang2022vision}. For alignment-based VLP models, we select CLIP\textsubscript{ViT} and CLIP\textsubscript{CNN}~\cite{radford2021learning} for evaluation.

\noindent\textbf{Adversarial attack settings.}
Following \cite{lu2023set}, we adopt BERT-Attack~\cite{li2020bert}, setting the text perturbation budget $\epsilon_t=1$ and using a vocabulary $W$ containing 10 candidate words for the text adversarial attack. For the image adversarial attack, we adopt the $l_\infty$ norm constraint with a perturbation budget of $\epsilon_v=8/255$.
We compare the proposed SADCA with several SOTA methods, including PGD~\cite{madry2018towards}, Bert-Attack~\cite{li2020bert}, Co-Attack~\cite{zhang2022towards}, SGA~\cite{lu2023set}, DRA~\cite{gao2024boosting}, and SA-AET~\cite{jia2024semantic}. 
For the original versions of these methods, we follow their default settings, configuring the image adversarial attack with an iteration number $J=10$.
Furthermore, we construct enhanced variants of SGA and SA-AET by combining them with the input transformation method SIA~\cite{wang2023structure}, resulting in SGA(LI)+SIA and SA-AET(LI)+SIA. These enhanced versions use a larger number of iterations $J=50$ to match the iteration budget of SADCA. 
For the proposed SADCA, we set the step size $\alpha=2/255$, momentum factor $\mu=1.0$, dynamic interaction iterations $I=5$, image attack iterations $J=10$, the number of negative samples $K=20$, the weighting factor for negative samples $\lambda=0.2$, and semantic-augmentation number $S=10$. 
All experiments are conducted on a single NVIDIA RTX 3090 GPU.

\noindent\textbf{Evaluation metrics.}
For image-text retrieval task , we use Attack Success Rate (ASR) at Rank-1 (R@1) as the evaluation metric for adversarial transferability. ASR represents the proportion of successfully attacked samples among all generated adversarial examples. A higher ASR indicates a more effective attack with greater transferability.
For visual grounding and image captioning tasks, we compare the performance of target models before and after attacks.

\subsection{Experimental Results}
\noindent\textbf{Cross-model adversarial transferability.}
We evaluate cross-model transferability on the ITR task using four representative VLP models: ALBEF, TCL, CLIP\textsubscript{ViT}, and CLIP\textsubscript{CNN}. 
Each model serves as the source model to generate adversarial examples, which are then transferred to the other three models. 
As shown in Table~\ref{cross-model}, SADCA achieves the highest average ASR in both TR and IR tasks, demonstrating strong generalization across architectures and training paradigms. 
For instance, when transferring from ALBEF to CLIP\textsubscript{CNN}, SADCA surpasses SA-AET(LI)+SIA by 9.19\% (TR) and 5.7\% (IR). 
Similarly, from CLIP\textsubscript{CNN} to CLIP\textsubscript{ViT}, it outperforms the second-best method by 7.61\% and 4.89\%, respectively. 
This indicates that the dynamic contrastive interaction and semantic augmentation mechanisms effectively promote semantic divergence, enabling adversarial examples to explore a wider range of attack directions within the semantic space.
Additional results on the MSCOCO dataset are provided in the supplementary material.

\begin{figure}[t]
\centering
\includegraphics[width=0.99\columnwidth]{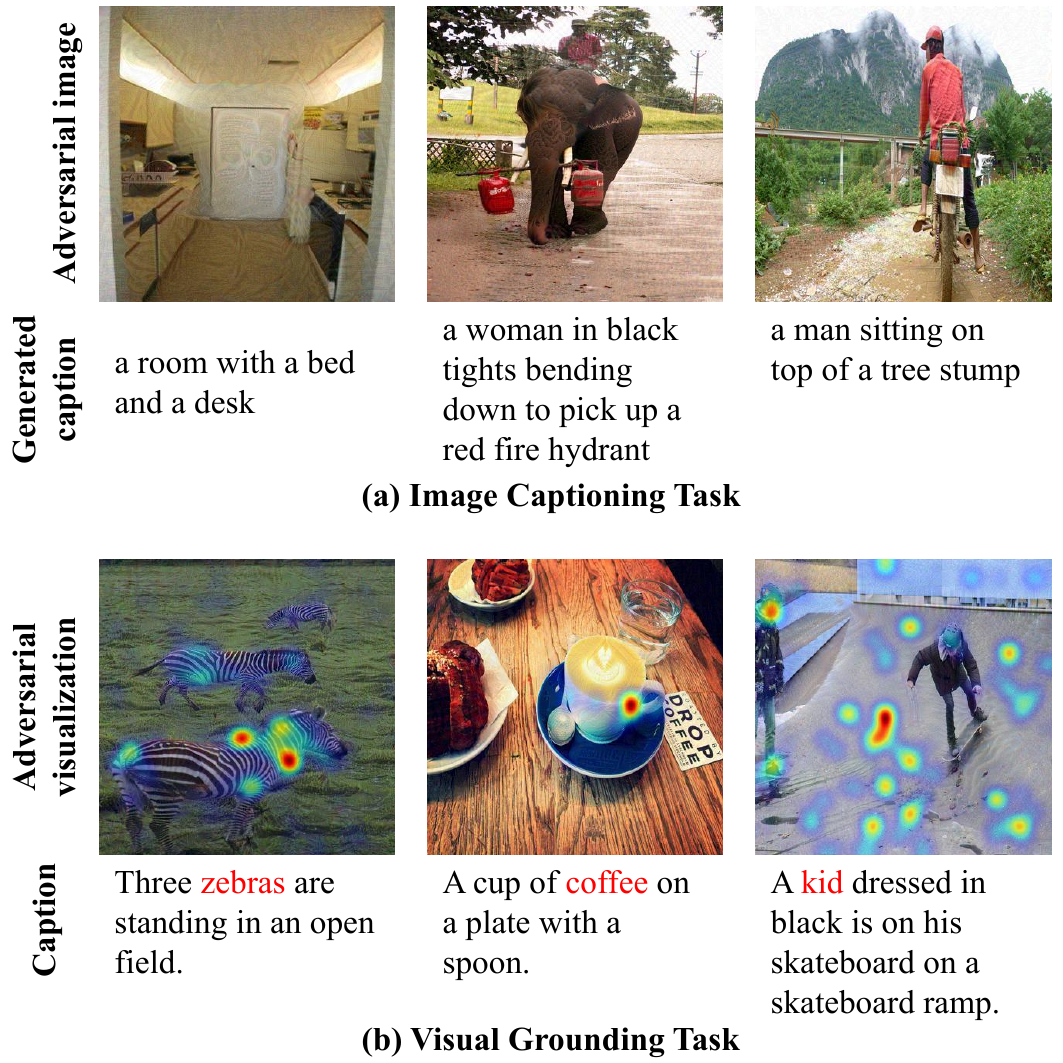}
\caption{Visualization on Image Captioning and Visual Grounding Tasks.}
\label{vis-task}
\end{figure}

\noindent\textbf{Cross-task adversarial transferability.}
In addition to evaluating the transferability of multimodal adversarial examples across different VLP models, we also conduct experiments to assess the effectiveness of the proposed SADCA in transferring across different vision-and-language (V+L) tasks. Specifically, we generate adversarial examples from the ITR task and evaluate their impact on the VG and IC tasks.
For IC task, we use the ALBEF model pretrained on the ITR task to generate adversarial images. These adversarial images are then fed into the BLIP~\cite{li2022blip} model to produce image captions.
For VG task, we apply the ALBEF model pretrained on the ITR task to generate adversarial images. Meanwhile, we employ the same model pretrained on VG task to localize the image regions.
As shown in the results of Table \ref{cross-task}, the multimodal adversarial examples exhibit strong cross-task adversarial transferability, successfully affecting performance on both the VG and IC tasks. Compared to other SOTA methods, the adversarial examples generated by SADCA lead to significantly larger performance drops in VG and IC metrics. This demonstrates the superior effectiveness of SADCA in cross-task adversarial transferability, consistently outperforming all other methods.
Figure \ref{vis-task} illustrates the results of image captioning and visual grounding on adversarial images generated by SADCA. As shown, both the generated captions and the localized regions deviate significantly from the expected outputs, indicating the effectiveness of SADCA.

\noindent\textbf{Adversarial transferability on LVLMs.}
With the increasing scale of data and model parameters, Large Vision-Language Models (LVLMs) have become increasingly popular. 
Therefore, we evaluate the performance of SADCA on a range of LVLMs, including open-source models such as LLaVA~\cite{liu2024improved}, Qwen3-VL~\cite{yang2025qwen3}, InternVL3~\cite{zhu2025internvl3}, and Qwen2.5-VL~\cite{bai2025qwen2}, as well as closed-source models including GPT-5, GPT-4o-mini, Gemini-2.0, and Claude-4.5.
Considering the differences between traditional VLP models and LVLM tasks, we redesign the experimental setup. 
Specifically, we employ a binary decision template (\emph{“Does the picture depict that ‘...’? Only answer Yes or No.”}) to construct adversarial text prompts, which are combined with adversarial images as model inputs.
The LVLMs are then prompted to make binary predictions under adversarial conditions.
As shown in Table~\ref{LVLMs}, SADCA consistently achieves the highest ASR across all models, revealing strong adversarial transferability.
Both open-source and closed-source LVLMs exhibit poor robustness when confronted with multimodal adversarial examples.
This observation highlights the importance of ensuring robustness against adversarial threats, which is as crucial as improving accuracy and overall performance in the development of LVLMs.

\begin{figure}[t]
\centering
\includegraphics[width=0.99\columnwidth]{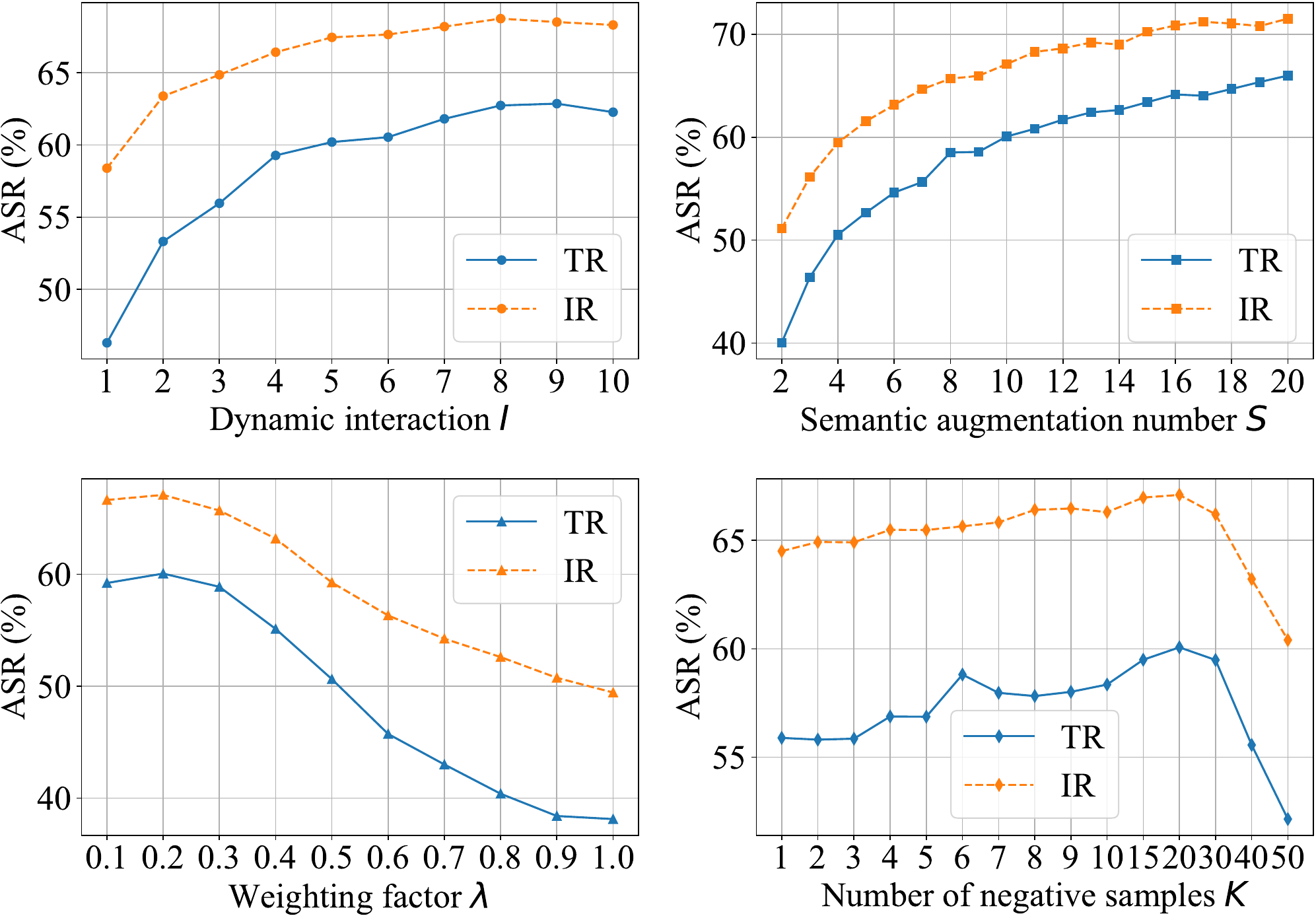}
\caption{The ASR (\%) with different parameters, including the dynamic interaction number $I$, the number of semantic augmentations $S$, the negative samples number $K$, and the weighting factor $\lambda$. The adversarial examples are generated on the Flickr30K dataset using CLIP\textsubscript{CNN} as the source model and are evaluated on other black-box models.}
\label{vis-parameter}
\end{figure}

\subsection{Parameter Analysis}
We conduct parameter experiments on the hyperparameters of SADCA, including the number of dynamic interaction iteration $I$, the number of semantic augmentations $S$, the number of negative samples $K$, and the weighting factor $\lambda$ for negative guidance. 
As shown in Figure \ref{vis-parameter}, increasing the number of dynamic interaction leads to a more thorough disruption of cross-modal interactions, resulting in stronger attack performance, though the effect tends to plateau at higher values.
A larger $S$ improves transferability by enriching semantic diversity, with $S=10$ offering a good trade-off between performance and efficiency.
Regarding the weighting factor $\lambda$, since negative samples primarily serve as a repulsive force in guiding the adversarial direction, smaller values of $\lambda$ tend to yield better performance. 
For $K$, a moderate value (\textit{e.g.}, $K=20$) balances effective guidance and noise suppression.

\begin{table}[t!]
\caption{Ablation study for negative sample selection strategy. The adversarial examples are generated by CLIP\textsubscript{CNN}.}
\centering
\setlength{\tabcolsep}{2pt}
\resizebox{1.0\linewidth}{!}{
\begin{tabular}{c|cc|cc|cc} 
\midrule
\multirow{2}{*}{\textbf{}}        & \multicolumn{2}{c|}{\textbf{ALBEF}} & \multicolumn{2}{c|}{\textbf{TCL}} & \multicolumn{2}{c}{\textbf{CLIPViT}}  \\
                                        & TR R@1         & IR R@1             & TR R@1         & IR R@1           & TR R@1         & IR R@1               \\ 
\midrule
Strategy (1)               & 44.00 & 56.13 & 49.43 & 59.33 & 74.11 & 77.51                \\
Strategy (2)            & 41.81 & 54.58 & 47.84 & 58.17 & 73.13 & 76.80                \\
Strategy (3)                & 44.53 & \textbf{58.40}  & 50.16 & 60.38 & 75.58 & \textbf{79.22}                \\
Strategy (4) & \textbf{45.15} & 57.48     & \textbf{51.42} & \textbf{60.43}   & \textbf{75.71} & 79.06       \\
\midrule
\end{tabular}
}
\label{Negative}
\end{table}

\begin{figure}[t]
\centering
\includegraphics[width=0.99\columnwidth]{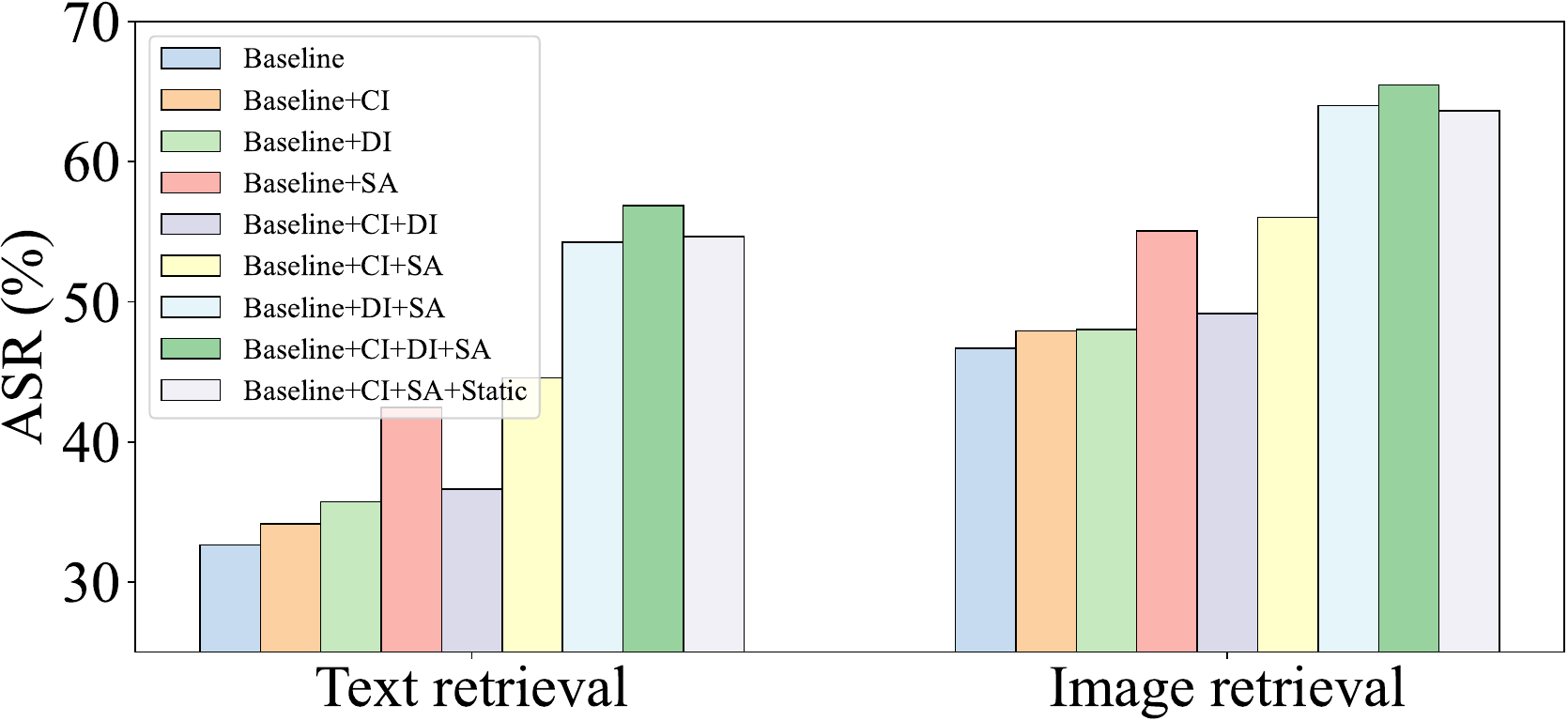}
\caption{Ablation study for different modules of SADCA. The adversarial examples are generated on the Flickr30K dataset using CLIP\textsubscript{CNN} as the source model and are evaluated on other black-box models.}
\label{vis-ablation}
\end{figure}

\subsection{Ablation Study}
\noindent\textbf{Impact of negative sample selection strategy.}
We explore several strategies for selecting negative samples:
(1) selecting the $K$ samples with the highest feature similarity to the positive sample;
(2) selecting the $K$ samples with the lowest feature similarity; and
(3) selecting the $K$ samples with intermediate similarity based on a ranked similarity list;
(4) randomly selecting $K$ negative samples.
As shown in Table \ref{Negative}, random selection achieves the best performance due to its higher sample diversity, which enhances generalization and transferability.
In contrast, selecting the most similar negatives leads to overfitting to local decision boundaries, while the least similar ones provide weak gradients. Intermediate negatives strike a balance but still lack sufficient diversity for strong generalization.

\noindent\textbf{Impact of different modules.}
We conduct ablation studies on the core components of SADCA, including Contrastive Interaction (CI), Dynamic Interaction (DI), and the Semantic Augmentation module (SA). 
As shown in Figure \ref{vis-ablation}, each component individually contributes to improved attack performance, while the SADCA with all components achieves the best overall results.
Additionally, we replace the dynamic interaction with a static interaction using the same iteration number. 
The observed performance drop under static interaction highlights the effectiveness of dynamic interaction in disrupting cross-modal alignment.
Notably, when the dynamic interaction and semantic augmentation modules are combined, a significant performance boost is achieved, indicating that their synergy jointly expands the coverage and disruption depth of adversarial perturbations in the semantic space.

\section{Conclusion}
In this paper, we focus on highly transferable vision-language attacks and propose SADCA, a novel method designed to enhance the adversarial transferability of multimodal adversarial examples. SADCA introduces a dynamic contrastive interaction mechanism to progressively disrupt cross-modal alignment and employs semantic-augmented transformations to increase the semantic diversity of adversarial examples. Extensive experiments conducted across various datasets and models demonstrate the effectiveness of SADCA, which consistently achieves superior transferability and outperforms SOTA methods.

\section{Acknowledgement}
This work was supported in part by the National Natural Science Foundation of China (62332008, 62576152, 62336004), the Basic Research Program of Jiangsu (BK20250104), the Fundamental Research Funds for the Central Universities (JUSRP202504007), and Leverhulme Trust Emeritus Fellowship EM-2025-06-09.

{
    \small
    \bibliographystyle{ieeenat_fullname}
    \bibliography{main}
}

\clearpage
\setcounter{page}{1}
\maketitlesupplementary

In this paper, we provide an illustration of the semantic augmentation module, experimental results on the MSCOCO dataset, , an evaluation of the effectiveness of the semantic augmentation module, and additional visualizations of the adversarial examples.
Furthermore, we also report the attack success rate at Rank-1 (R@1), Rank-5 (R@5), and Rank-10 (R@10) on the Flickr30K dataset.

\section{Semantic Augmentation Module}
\begin{figure}[t!]
\centering
\includegraphics[width=0.99\columnwidth]{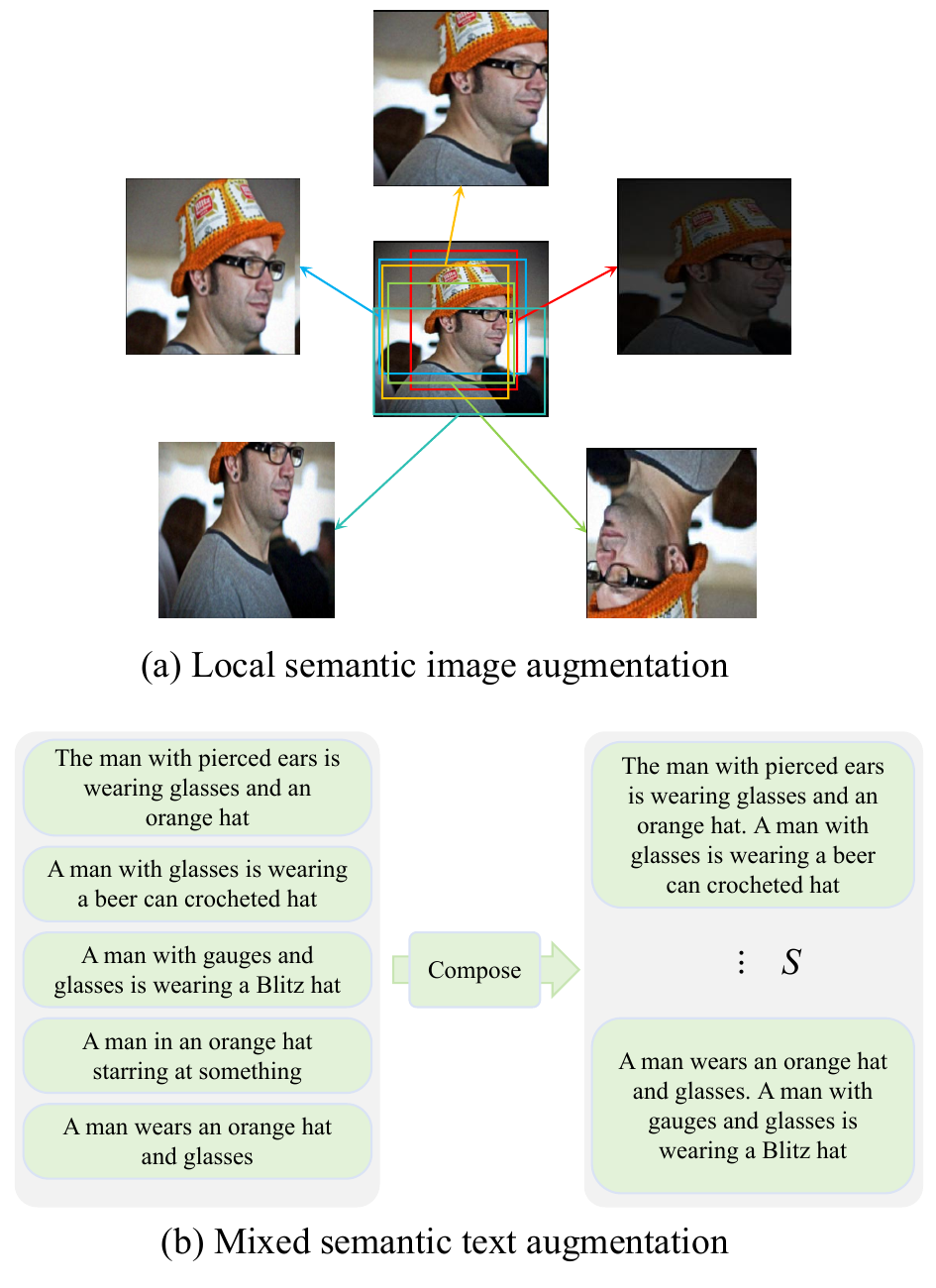}
\caption{Semantic Augmentation Module.}
\label{illustration}
\end{figure}

Figure \ref{illustration} illustrates the Semantic Augmentation Module. For image local semantic augmentation, the module obtains enlarged local regions by randomly cropping and resizing parts of the image, followed by the application of random image augmentation functions to diversify the local semantic content. For text mixed semantic augmentation, the module randomly selects and concatenates pairs of text samples from the text pool to generate a new augmented text set. This approach combines multiple textual descriptions to form broader semantic representations, thereby enhancing semantic diversity.

\section{Experimental Results on MSCOCO dataset}
We conduct comparative experiments on the ITR task using four widely adopted VLP models: ALBEF, TCL, CLIP\textsubscript{ViT} and CLIP\textsubscript{CNN}. 
Specifically, each VLP model is used as a source model to generate multimodal adversarial examples, which are then evaluated on the remaining three models to assess cross-model transferability. 
Table~\ref{cross-model} presents the comparative results on the MSCOCO dataset. As shown, SADCA consistently achieves the highest average black-box ASR in both TR and IR tasks. 
These results demonstrate the effectiveness of SADCA in significantly enhancing the cross-model transferability of multimodal adversarial examples.

\begin{table*}[t]
\caption{Comparison with SOTA methods on the image-text retrieval (ITR) task on the MSCOCO dataset. The "Source" column indicates the VLP model used to generate the multimodal adversarial examples. For both image retrieval (IR) and text retrieval (TR), we report the ASR (\%) at Rank-1 (R@1). The "Average" represents the average ASR on the black-box VLP models.}
\centering
\small
\setlength{\tabcolsep}{3pt}
\begin{tabular}{c|c|cc|cc|cc|cc|cc}
\midrule
\multirow{2}{*}{\textbf{Source}} & \multirow{2}{*}{\textbf{Attack}} & \multicolumn{2}{c|}{\textbf{ALBEF}} & \multicolumn{2}{c|}{\textbf{TCL}} & \multicolumn{2}{c|}{\textbf{CLIP\textsubscript{ViT}}} & \multicolumn{2}{c|}{\textbf{CLIP\textsubscript{CNN}}}  & \multicolumn{2}{c}{\textbf{Average}} \\
 &  & TR R@1 & IR R@1 & TR R@1 & IR R@1 & TR R@1 & IR R@1 & TR R@1 & IR R@1 & TR R@1 & IR R@1 \\
\midrule
\multirow{9}{*}{\textbf{ALBEF}} 
& PGD & 94.35 & 93.26 & 34.15 & 36.86 & 21.71 & 27.06 & 23.83 & 30.96 & 26.56 & 31.63 \\
& BERT-Attack & 24.39 & 36.13 & 24.34 & 33.39 & 44.94 & 52.28 & 47.73 & 54.75 & 39.00 & 46.81 \\
& Co-Attack & 94.95 & 97.87 & 65.22 & 72.41 & 55.28 & 62.33 & 56.68 & 66.45 & 59.06 & 67.06 \\
& SGA & 99.95 & 99.94 & 87.46 & 88.17 & 63.72 & 69.71 & 63.91 & 70.78 & 71.70 & 76.22 \\
& SGA(LI)+SIA & 100 & 100 & 98.49 & 98.01 & 75.96 & 79.83 & 76.75 & 81.21 & 83.73 & 86.35 \\
& DRA & 99.90 & 99.93 & 88.81 & 90.06 & 69.25 & 75.31 & 68.53 & 75.09 & 75.53 & 80.15 \\
& SA-AET & 100 & 99.99 & 97.28 & 96.88 & 76.57 & 80.24 & 76.17 & 80.64 & 85.92 & 83.34 \\
& SA-AET(LI)+SIA & 100 & 100 & \textbf{99.66} & \textbf{99.50} & 91.87 & 92.69 & 90.56 & 92.84 & 94.03 & 95.01 \\
& \cellcolor{gray!15}SADCA (ours) & \cellcolor{gray!15}\textbf{100} & \cellcolor{gray!15}\textbf{100} & \cellcolor{gray!15}98.35 & \cellcolor{gray!15}97.99 & \cellcolor{gray!15}\textbf{93.10} & \cellcolor{gray!15}\textbf{94.01} & \cellcolor{gray!15}\textbf{93.44} & \cellcolor{gray!15}\textbf{95.01} &
\cellcolor{gray!15}\textbf{94.96} & \cellcolor{gray!15}\textbf{95.67} \\
\midrule
\multirow{9}{*}{\textbf{TCL}} 
& PGD & 40.81 & 44.09 & 98.54 & 98.20 & 21.79 & 26.92 & 24.97 & 32.17 & 29.19 & 34.39 \\
& BERT-Attack & 35.32 & 45.92 & 38.54 & 48.48 & 51.09 & 58.80 & 52.23 & 61.26 & 46.21 & 55.33 \\
& Co-Attack & 49.84 & 60.36 & 91.68 & 95.48 & 32.64 & 42.69 & 32.06 & 47.82 & 38.85 & 50.29 \\
& SGA & 92.70 & 92.99 & 100 & 100 & 59.79 & 65.31 & 60.52 & 67.34 & 71.00 & 75.21 \\
& SGA(LI)+SIA & 99.94 & 99.43 & 100 & 100 & 74.55 & 78.94 & 78.79 & 82.62 & 84.43 & 86.99 \\
& DRA & 94.72 & 95.89 & 100 & 100 & 70.51 & 74.95 & 70.29 & 76.99 & 78.51 & 82.61 \\
& SA-AET & 97.78 & 98.08 & 100 & 99.99 & 76.12 & 79.74 & 75.89 & 80.92 & 83.26 & 86.25 \\
& SA-AET(LI)+SIA & \textbf{99.85} & \textbf{99.78} & 100 & 100 & 90.92 & 92.92 & 92.77 & 94.49 & 94.51 & 95.73 \\
& \cellcolor{gray!15}SADCA (ours) & \cellcolor{gray!15}99.59 & \cellcolor{gray!15}99.46 & \cellcolor{gray!15}\textbf{100} & \cellcolor{gray!15}\textbf{100} & \cellcolor{gray!15}\textbf{93.52} & \cellcolor{gray!15}\textbf{94.39} & \cellcolor{gray!15}\textbf{95.59} & \cellcolor{gray!15}\textbf{96.62}
& \cellcolor{gray!15}\textbf{96.23} & \cellcolor{gray!15}\textbf{96.82}\\
\midrule
\multirow{9}{*}{\textbf{CLIP\textsubscript{ViT}}}
& PGD & 10.26 & 13.69 & 12.72 & 15.81 & 82.91 & 90.51 & 21.62 & 28.78 & 14.87 & 19.43 \\
& BERT-Attack & 20.34 & 29.74 & 21.08 & 29.61 & 45.06 & 51.68 & 44.54 & 53.72 & 28.65 & 37.69 \\
& Co-Attack & 26.35 & 36.69 & 28.23 & 38.42 & 88.78 & 96.72 & 47.36 & 58.45 & 33.98 & 44.52 \\
& SGA & 43.75 & 51.08 & 44.05 & 51.02 & 100 & 100 & 70.66 & 75.58 & 52.82 & 59.23 \\
& SGA(LI)+SIA & 62.49 & 65.52 & 64.07 & 65.21 & 100 & 100 & 93.09 & 94.80 & 73.22 & 75.18 \\
& DRA & 52.69 & 61.50 & 51.88 & 61.06 & 100 & 100 & 80.18 & 84.11 & 61.58 & 68.89 \\
& SA-AET & 57.64 & 66.88 & 57.30 & 65.16 & 100 & 100 & 83.98 & 86.72 & 66.31 & 72.92  \\
& SA-AET(LI)+SIA & 86.23 & 87.47 & 85.42 & 86.16 & 100 & 100 & 97.59 & 97.96 & 89.75 & 90.53 \\
& \cellcolor{gray!15}SADCA (ours) & \cellcolor{gray!15}\textbf{90.79} & \cellcolor{gray!15}\textbf{91.09} & \cellcolor{gray!15}\textbf{88.46} & \cellcolor{gray!15}\textbf{87.90} & \cellcolor{gray!15}\textbf{100} & \cellcolor{gray!15}\textbf{100} & \cellcolor{gray!15}\textbf{99.51} & \cellcolor{gray!15}\textbf{99.53} & \cellcolor{gray!15}\textbf{92.92} & \cellcolor{gray!15}\textbf{92.84}\\
\midrule
\multirow{9}{*}{\textbf{CLIP\textsubscript{CNN}}} 
& PGD & 8.38 & 12.73 & 11.90 & 15.68 & 13.66 & 20.62 & 92.68 & 94.71 & 11.31 & 16.34 \\
& BERT-Attack & 23.38 & 34.64 & 24.58 & 29.61 & 51.28 & 57.49 & 54.43 & 62.17 & 33.08 & 40.58 \\
& Co-Attack & 29.49 & 41.50 & 31.83 & 43.44 & 53.15 & 60.15 & 97.79 & 98.54 & 38.16 & 48.36 \\
& SGA & 36.94 & 46.79 & 38.81 & 48.90 & 62.19 & 67.70 & 97.79 & 98.54 & 45.98 & 54.46 \\
& SGA(LI)+SIA & 37.15 & 45.49 & 39.81 & 48.26 & 65.32 & 72.56 & 100 & 100 & 47.43 & 55.44

 \\
& DRA & 41.40 & 52.25 & 43.62 & 54.15 & 70.43 & 74.14 & 99.80 & 99.92 & 51.82 & 60.18 \\
& SA-AET & 43.62 & 55.19 & 47.01 & 57.39 & 73.67 & 76.90 & 100 & 99.92 & 54.77 & 63.16 \\
& SA-AET(LI)+SIA & 55.35 & 63.75 & 58.49 & 65.66 & 85.5 & 87.41 & 100 & 100 & 66.45 & 72.27 \\
& \cellcolor{gray!15}SADCA (ours) & \cellcolor{gray!15}\textbf{62.90} & \cellcolor{gray!15}\textbf{69.64} & \cellcolor{gray!15}\textbf{63.78} & \cellcolor{gray!15}\textbf{69.48} & \cellcolor{gray!15}\textbf{88.45} & \cellcolor{gray!15}\textbf{91.16} & \cellcolor{gray!15}\textbf{100} & \cellcolor{gray!15}\textbf{100} & \cellcolor{gray!15}\textbf{71.71} & \cellcolor{gray!15}\textbf{76.76} \\
\midrule
\end{tabular}
\label{cross-model}
\end{table*}

\begin{table*}[t]
\caption{Ablation Study for Semantic Augmentation Module. }
\centering
\small
\begin{tabular}{c|c|cc|cc|cc|cc}
\midrule
\multirow{2}{*}{\textbf{Source}} & \multirow{2}{*}{\textbf{Attack}} & \multicolumn{2}{c|}{\textbf{ALBEF}} & \multicolumn{2}{c|}{\textbf{TCL}} & \multicolumn{2}{c|}{\textbf{CLIP\textsubscript{ViT}}} & \multicolumn{2}{c}{\textbf{CLIP\textsubscript{CNN}}}  \\
 &  & TR R@1 & IR R@1 & TR R@1 & IR R@1 & TR R@1 & IR R@1 & TR R@1 & IR R@1 \\
\midrule
\multirow{4}{*}{\textbf{ALBEF}} 
& DCI+DIM & 100 & 100 & 94.73 & 94.05 & 73.37 & 77.93 & 74.46 & 77.77  \\
& DCI+SIA & 100 & 100 & \textbf{99.58} & \textbf{99.57} & 70.55 & 74.58 & 73.18 & 78.66  \\
& DCI+BSR & 100 & 100 & 98.74 & 98.86 & 77.30 & 79.19 & 78.80 & 82.20  \\
& \cellcolor{gray!15}SADCA & \cellcolor{gray!15}\textbf{100} & \cellcolor{gray!15}\textbf{100} & \cellcolor{gray!15}98.52 & \cellcolor{gray!15}97.83 & \cellcolor{gray!15}\textbf{81.10} & \cellcolor{gray!15}\textbf{82.83} & \cellcolor{gray!15}\textbf{85.44} & \cellcolor{gray!15}\textbf{86.11}  \\
\midrule
\multirow{4}{*}{\textbf{TCL}} 
& DCI+DIM & 97.60 & 97.26 & 100 & 100 & 75.71 & 80.54 & 80.08 & 83.53  \\
& DCI+SIA & \textbf{100}  & \textbf{99.95} & 100 & 100 & 71.66 & 78.38 & 79.44 & 83.40   \\
& DCI+BSR & 100  & 99.86 & 100 & 100 & \textbf{79.97} & 82.89 & 86.39 & 88.30   \\
& \cellcolor{gray!15}SADCA & \cellcolor{gray!15}99.58 & \cellcolor{gray!15}99.56 & \cellcolor{gray!15}\textbf{100} & \cellcolor{gray!15}\textbf{100} & \cellcolor{gray!15}78.28 & \cellcolor{gray!15}\textbf{83.18} & \cellcolor{gray!15}\textbf{86.46} & \cellcolor{gray!15}\textbf{88.71}\\
\midrule
\multirow{4}{*}{\textbf{CLIP\textsubscript{ViT}}}
& DCI+DIM & 49.95 & 57.60 & 48.37 & 57.57 & 100   & 100   & 80.59 & 82.64   \\
& DCI+SIA & 77.69 & 80.24 & 79.45 & 81.98 & 100   & 100   & 96.93 & 96.64   \\
& DCI+BSR & 79.35 & 81.55 & 80.30 & 82.83 & 100   & 100   & 96.96 & 97.32   \\
& \cellcolor{gray!15}SADCA & \cellcolor{gray!15}\textbf{87.07} & \cellcolor{gray!15}\textbf{89.20} & \cellcolor{gray!15}\textbf{87.04} & \cellcolor{gray!15}\textbf{87.98} & \cellcolor{gray!15}\textbf{100} & \cellcolor{gray!15}\textbf{100} & \cellcolor{gray!15}\textbf{97.90} & \cellcolor{gray!15}\textbf{97.46} \\
\midrule
\multirow{4}{*}{\textbf{CLIP\textsubscript{CNN}}} 
& DCI+DIM & 23.77 & 40.34 & 29.5  & 43.33 & 56.69 & 65.01 & 100   & 100   \\
& DCI+SIA & 42.46 & 55.96 & 47.63 & 60.02 & 76.56 & 80.80 & 100   & 100   \\
& DCI+BSR & 37.02 & 51.50 & 44.26 & 55.79 & 69.20 & 75.58 & 100   & 100   \\
& \cellcolor{gray!15}SADCA & \cellcolor{gray!15}\textbf{49.43} & \cellcolor{gray!15}\textbf{60.55} & \cellcolor{gray!15}\textbf{55.53} & \cellcolor{gray!15}\textbf{63.19} & \cellcolor{gray!15}\textbf{77.18} & \cellcolor{gray!15}\textbf{79.57} & \cellcolor{gray!15}\textbf{100} & \cellcolor{gray!15}\textbf{100} \\
\midrule
\end{tabular}
\label{SAM}
\end{table*}

\begin{table*}[t!]
\caption{Comparison of attack costs. GPU memory usage, runtime, and attack performance when generating adversarial examples on the Flickr30K dataset using ALBEF as the surrogate model.}
\centering
\begin{tabular}{c|cccc} 
\midrule
Methods        & GPU Memory (GB) & Run Time (h) & TR R@1 & IR R@1  \\ 
\midrule
SGA            & 10.5           & 0.83        & 54.72  & 61.54   \\
SGA(LI)+SIA    & 13.0           & 3.95        & 67.99  & 72.07   \\
DRA            & 10.6           & 1.57        & 62.46  & 69.06   \\
SA-AET         & 10.5           & 2.12        & 69.74  & 75.17   \\
SA-AET(LI)+SIA & 13.5           & 11.08       & 83.85  & 86.12   \\
\rowcolor[rgb]{0.925,0.925,0.925} SADCA          & 13.3           & 4.40        & 88.35  & 88.92   \\
\midrule
\end{tabular}
\label{cost}
\end{table*}

\section{Ablation Study for Semantic Augmentation Module}
To validate the advantages of the Semantic Augmentation Module, we compare it with other input transformation methods. Specifically, we retain the Dynamic Contrastive Interaction (DCI) component and replace the Semantic Augmentation Module with three alternative input transformation methods: DIM~\cite{xie2019improving}, SIA~\cite{wang2023structure}, and BSR~\cite{wang2024boosting}. As shown in Table \ref{SAM}, SADCA equipped with the Semantic Augmentation Module consistently outperforms the alternatives in most cases. This demonstrates that the Semantic Augmentation Module is well-suited to VLP models, effectively enhancing the semantic diversity of inputs and thereby disrupting the alignment mechanisms within vision-language models more efficiently.

\section{A Comparison of Attack Cost}
Table \ref{cost} summarizes the GPU memory usage, runtime, and attack performance when generating adversarial examples using ALBEF as the surrogate model. SADCA achieves the strongest attack performance, reaching 88.35\% (TR R@1) and 88.92\% (IR R@1), outperforming all compared methods by a clear margin. Although its GPU memory consumption (13.3 GB) and runtime (4.40 h) are moderately higher than those of simpler baselines, they remain substantially lower than the most computationally expensive method, SA-AET(LI)+SIA, while delivering significantly better results. This demonstrates that SADCA offers the most favorable cost–performance balance, providing superior adversarial transferability with a reasonable computational overhead.

\section{More Results on Flickr30K Dataset}
Reporting only the ASR at R@1 (i.e., the correct image no longer appearing at the top rank) is insufficient for a comprehensive evaluation of the robustness of multimodal retrieval models, as the correct sample may merely be shifted to Rank-2 or Rank-3, which has limited impact on the overall system. In contrast, presenting the attack success rates at R@5 and R@10 provides insight into whether the attack truly disrupts a broader portion of the ranking structure, thereby offering a more complete assessment of the model’s vulnerability at the ranking level.
Accordingly, we report the ASR at Rank-1, Rank-5, and Rank-10 on the Flickr30K dataset in Table \ref{TR1510} and Table \ref{IR1510}. The results show that SADCA achieves strong attack performance across all metrics, indicating its effectiveness in substantially perturbing the retrieval ranking system and demonstrating its stronger attack capability.

\begin{table*}[t]
\caption{Comparison with SOTA methods on the image-text retrieval (ITR) task on the Flickr30K dataset. The "Source" column indicates the VLP model used to generate the multimodal adversarial examples. For text retrieval (TR), we report the ASR (\%) at Rank-1 (R@1), Rank-5 (R@5) and Rank-10 (R@10).}
\centering
\setlength{\tabcolsep}{2pt}
\resizebox{1.0\textwidth}{!}{
\begin{tabular}{c|c|ccc|ccc|ccc|ccc}
\midrule
\multirow{2}{*}{Source} & \multirow{2}{*}{Attack} & \multicolumn{3}{c|}{ALBEF} & \multicolumn{3}{c|}{TCL} & \multicolumn{3}{c|}{CLIP\textsubscript{ViT}} & \multicolumn{3}{c}{CLIP\textsubscript{CNN}} \\
& & TR R@1 & TR R@5 & TR R@10 & TR R@1 & TR R@5 & TR R@10 & TR R@1 & TR R@5 & TR R@10 & TR R@1 & TR R@5 & TR R@10 \\
\midrule
\multirow{10}{*}{ALBEF}
& SGA            & 99.90 & 99.70 & 99.70 & 87.88 & 77.79 & 71.74 & 36.69 & 19.83 & 12.40 & 39.59 & \textbf{21.88} & 14.83 \\
& SGA+SIA        & 99.79 & 99.50 & 99.20 & 94.10 & 88.74 & 85.07 & 47.48 & 29.60 & 21.75 & 53.77 & 34.46 & 24.82 \\
& SGA(LI)        & 100 & 100 & 100 & 84.19 & 73.17 & 67.43 & 4.36 & 15.47 & 10.06 & 36.53 & 19.77 & 12.36 \\
& SGA(LI)+SIA    & 100 & 100 & 100 & 99.37 & 98.29 & 97.80 & 49.82 & 32.40 & 23.68 & 54.79 & 35.31 & 28.01 \\
& DRA            & 99.90 & 99.70 & 99.70 & 91.57 & 81.31 & 75.95 & 46.26 & 25.44 & 18.60 & 49.55 & 29.49 & 21.22 \\
& SA-AET         & 99.90 & 99.80 & 99.80 & 96.42 & 92.36 & 89.98 & 55.58 & 34.48 & 26.32 & 57.22 & 39.64 & 28.53 \\
& SA-AET+SIA     & 99.58 & 99.30 & 98.80 & 95.21 & 90.01 & 85.97 & 64.54 & 41.64 & 34.04 & 66.16 & 44.82 & 36.35 \\
& SA-AET(LI)     & 100 & 100 & 100 & 98.63 & 97.29 & 96.09 & 60.61 & 38.11 & 30.18 & 62.20 & 41.12 & 32.65 \\
& SA-AET(LI)+SIA & 100 & 100 & 100 & \textbf{99.58} & \textbf{98.89} & \textbf{98.50} & 75.71 & 60.12 & 52.13 & 76.25 & 61.21 & 53.66 \\
& \cellcolor[rgb]{0.925,0.925,0.925}{SADCA}          & \cellcolor[rgb]{0.925,0.925,0.925}{\textbf{100}} & \cellcolor[rgb]{0.925,0.925,0.925}{\textbf{100}} & \cellcolor[rgb]{0.925,0.925,0.925}{\textbf{100}} & \cellcolor[rgb]{0.925,0.925,0.925}{98.52} & \cellcolor[rgb]{0.925,0.925,0.925}{96.28} & \cellcolor[rgb]{0.925,0.925,0.925}{94.79} & \cellcolor[rgb]{0.925,0.925,0.925}{\textbf{81.10}} & \cellcolor[rgb]{0.925,0.925,0.925}{\textbf{62.10}} & \cellcolor[rgb]{0.925,0.925,0.925}{\textbf{55.08}} & \cellcolor[rgb]{0.925,0.925,0.925}{\textbf{85.44}} & \cellcolor[rgb]{0.925,0.925,0.925}{\textbf{72.09}} & \cellcolor[rgb]{0.925,0.925,0.925}{\textbf{65.91}} \\
\midrule
\multirow{10}{*}{TCL}
& SGA            & 92.40 & 87.07 & 85.40 & 100 & 100 & 99.90 & 36.81 & 18.59 & 13.11 & 41.89 & 22.73 & 14.93 \\
& SGA+SIA        & 97.18 & 94.19 & 91.90 & 100 & 99.70 & 99.60 & 53.01 & 33.13 & 27.03 & 58.49 & 41.33 & 32.34 \\
& SGA(LI)        & 84.46 & 76.25 & 71.50 & 100 & 100 & 100 & 31.17 & 15.06 & 9.86 & 36.78 & 20.08 & 14.01 \\
& SGA(LI)+SIA    & 99.90 & 99.70 & 99.60 & 100 & 99.90 & 99.90 & 52.87 & 33.85 & 26.22 & 62.45 & 44.19 & 36.87 \\
& DRA            & 95.20 & 91.28 & 88.00 & 100 & 100 & 99.80 & 47.24 & 26.48 & 18.90 & 52.23 & 30.76 & 22.97 \\
& SA-AET         & 98.85 & 96.79 & 95.50 & 100 & 100 & 100 & 56.20 & 34.68 & 25.91 & 59.77 & 39.22 & 30.07 \\
& SA-AET+SIA     & 97.08 & 93.79 & 92.60 & 99.79 & 99.60 & 99.20 & 67.36 & 49.53 & 40.65 & 70.11 & 54.65 & 46.14 \\
& SA-AET(LI)     & 99.37 & 98.60 & 98.20 & 100 & 100 & 100 & 56.20 & 37.49 & 28.05 & 62.32 & 42.81 & 32.75 \\
& SA-AET(LI)+SIA & \textbf{99.95} & \textbf{99.90} & \textbf{99.80} & 100 & 100 & 100 & 77.04 & 65.52 & 57.52 & 80.20 & 67.86 & 61.59 \\
& \cellcolor[rgb]{0.925,0.925,0.925}{SADCA}          & \cellcolor[rgb]{0.925,0.925,0.925}{99.58} & \cellcolor[rgb]{0.925,0.925,0.925}{99.20} & \cellcolor[rgb]{0.925,0.925,0.925}{99.00} & \cellcolor[rgb]{0.925,0.925,0.925}{\textbf{100}} & \cellcolor[rgb]{0.925,0.925,0.925}{\textbf{100}} & \cellcolor[rgb]{0.925,0.925,0.925}{\textbf{100}} & \cellcolor[rgb]{0.925,0.925,0.925}{\textbf{78.28}} & \cellcolor[rgb]{0.925,0.925,0.925}{\textbf{65.94}} & \cellcolor[rgb]{0.925,0.925,0.925}{\textbf{58.64}} & \cellcolor[rgb]{0.925,0.925,0.925}{\textbf{86.46}} & \cellcolor[rgb]{0.925,0.925,0.925}{\textbf{76.74}} & \cellcolor[rgb]{0.925,0.925,0.925}{\textbf{69.93}} \\
\midrule
\multirow{10}{*}{CLIP\textsubscript{ViT}}
& SGA            & 22.42 & 9.02 & 5.20 & 25.08 & 9.55 & 6.01 & 100 & 100 & 99.90 & 53.26 & 33.83 & 25.33 \\
& SGA+SIA        & 44.94 & 25.65 & 19.60 & 47.10 & 27.24 & 20.44 & 99.88 & 99.48 & 99.09 & 81.48 & 65.12 & 55.10 \\
& SGA(LI)        & 21.17 & 8.92 & 5.60 & 23.71 & 8.74 & 5.51 & 100 & 100 & 100 & 49.43 & 31.40 & 21.52 \\
& SGA(LI)+SIA    & 54.33 & 34.37 & 27.90 & 57.43 & 36.88 & 30.76 & 100 & 100 & 100 & 89.02 & 80.66 & 72.61 \\
& DRA            & 27.84 & 12.73 & 8.10 & 27.82 & 12.46 & 7.52 & 100 & 100 & 100 & 64.88 & 42.49 & 33.47 \\
& SA-AET         & 36.60 & 20.64 & 16.00 & 39.20 & 20.30 & 14.33 & 100 & 100 & 100 & 71.01 & 50.11 & 41.50 \\
& SA-AET+SIA     & 54.54 & 36.17 & 30.20 & 58.59 & 38.79 & 31.96 & 99.63 & 99.48 & 98.98 & 83.91 & 69.34 & 62.00 \\
& SA-AET(LI)     & 45.57 & 28.76 & 21.80 & 46.89 & 27.44 & 21.14 & 100 & 100 & 100 & 78.67 & 60.89 & 52.01 \\
& SA-AET(LI)+SIA & 79.04 & 64.33 & 58.40 & 79.35 & 66.33 & 59.72 & 100 & 100 & 100 & 94.76 & 89.32 & 85.79 \\
& \cellcolor[rgb]{0.925,0.925,0.925}{SADCA}          & \cellcolor[rgb]{0.925,0.925,0.925}{\textbf{87.07}} & \cellcolor[rgb]{0.925,0.925,0.925}{\textbf{74.85}} & \cellcolor[rgb]{0.925,0.925,0.925}{\textbf{68.40}} & \cellcolor[rgb]{0.925,0.925,0.925}{\textbf{87.04}} & \cellcolor[rgb]{0.925,0.925,0.925}{\textbf{74.07}} & \cellcolor[rgb]{0.925,0.925,0.925}{\textbf{66.93}} & \cellcolor[rgb]{0.925,0.925,0.925}{\textbf{100}} & \cellcolor[rgb]{0.925,0.925,0.925}{\textbf{100}} & \cellcolor[rgb]{0.925,0.925,0.925}{\textbf{100}} & \cellcolor[rgb]{0.925,0.925,0.925}{\textbf{97.90}} & \cellcolor[rgb]{0.925,0.925,0.925}{\textbf{92.81}} & \cellcolor[rgb]{0.925,0.925,0.925}{\textbf{89.91}} \\
\midrule
\multirow{10}{*}{CLIP\textsubscript{CNN}}
& SGA            & 15.64 & 5.61 & 3.00 & 18.02 & 6.03 & 2.91 & 39.92 & 20.15 & 13.52 & 99.87 & 99.47 & 99.07 \\
& SGA+SIA        & 19.29 & 6.11 & 4.00 & 22.76 & 8.34 & 5.21 & 46.50 & 23.88 & 16.16 & 99.62 & 97.99 & 96.81 \\
& SGA(LI)        & 15.51 & 5.61 & 2.50 & 17.70 & 6.13 & 3.51 & 38.16 & 18.38 & 12.30 & 99.87 & 99.89 & 99.90 \\
& SGA(LI)+SIA    & 20.96 & 9.02 & 5.40 & 22.55 & 11.06 & 7.21 & 47.24 & 26.27 & 19.41 & 100 & 99.68 & 99.38 \\
& DRA            & 19.50 & 6.31 & 3.70 & 21.60 & 7.54 & 3.81 & 48.47 & 99.47 & 17.89 & 99.87 & 25.13 & 99.38 \\
& SA-AET         & 23.98 & 9.22 & 6.00 & 27.29 & 10.45 & 6.61 & 54.11 & 33.33 & 24.39 & 100 & 100 & 99.90 \\
& SA-AET+SIA     & 26.80 & 12.63 & 8.70 & 31.61 & 13.77 & 8.52 & 57.18 & 35.62 & 25.20 & 99.49 & 98.31 & 96.70 \\
& SA-AET(LI)     & 31.07 & 15.03 & 12.00 & 33.19 & 15.08 & 11.62 & 61.72 & 43.09 & 34.45 & 100 & 100 & 100 \\
& SA-AET(LI)+SIA & 38.69 & 20.74 & 17.10 & 44.89 & 25.13 & 19.74 & 69.57 & 52.34 & 42.89 & 100 & 99.79 & 99.79 \\
& \cellcolor[rgb]{0.925,0.925,0.925}{SADCA}          & \cellcolor[rgb]{0.925,0.925,0.925}{\textbf{49.43}} & \cellcolor[rgb]{0.925,0.925,0.925}{\textbf{26.05}} & \cellcolor[rgb]{0.925,0.925,0.925}{\textbf{20.90}} & \cellcolor[rgb]{0.925,0.925,0.925}{\textbf{55.53}} & \cellcolor[rgb]{0.925,0.925,0.925}{\textbf{30.15}} & \cellcolor[rgb]{0.925,0.925,0.925}{\textbf{23.45}} & \cellcolor[rgb]{0.925,0.925,0.925}{\textbf{77.18}} & \cellcolor[rgb]{0.925,0.925,0.925}{\textbf{56.39}} & \cellcolor[rgb]{0.925,0.925,0.925}{\textbf{48.48}} & \cellcolor[rgb]{0.925,0.925,0.925}{\textbf{100}} & \cellcolor[rgb]{0.925,0.925,0.925}{\textbf{100}} & \cellcolor[rgb]{0.925,0.925,0.925}{\textbf{100}} \\
\midrule
\end{tabular}
}
\label{TR1510}
\end{table*}

\begin{table*}[t]
\caption{Comparison with SOTA methods on the image-text retrieval (ITR) task on the Flickr30K dataset. The "Source" column indicates the VLP model used to generate the multimodal adversarial examples. For image retrieval (IR), we report the ASR (\%) at Rank-1 (R@1), Rank-5 (R@5) and Rank-10 (R@10).}
\centering
\setlength{\tabcolsep}{2pt}
\resizebox{1.0\textwidth}{!}{
\begin{tabular}{c|c|ccc|ccc|ccc|ccc}
\midrule
\multirow{2}{*}{Source}&\multirow{2}{*}{Attack}&\multicolumn{3}{c|}{ALBEF}&\multicolumn{3}{c|}{TCL}&\multicolumn{3}{c|}{CLIP\textsubscript{ViT}}&\multicolumn{3}{c}{CLIP\textsubscript{CNN}}\\
& &IR R@1&IR R@5&IR R@10&IR R@1&IR R@5&IR R@10&IR R@1&IR R@5&IR R@10&IR R@1&IR R@5&IR R@10\\
\midrule
\multirow{10}{*}{ALBEF}
&SGA&99.93&99.92&99.90&88.05&77.65&71.32&46.78&29.29&22.18&49.78&32.70&24.80\\
&SGA+SIA&99.79&99.41&99.15&94.21&88.11&83.33&57.38&38.68&30.88&61.23&43.96&35.19\\
&SGA(LI)&100&100&100&84.50&72.94&66.21&41.04&23.41&17.60&45.18&26.95&20.71\\
&SGA(LI)+SIA&100&100&100&99.02&97.40&96.28&56.19&38.03&30.77&60.99&43.86&34.63\\
&DRA&99.93&99.86&99.82&91.36&82.24&77.03&56.80&38.61&29.79&59.01&41.65&33.52\\
&SA-AET&99.98&100&99.98&96.02&92.02&89.05&63.89&45.88&36.52&65.59&48.84&40.03\\
&SA-AET+SIA&99.72&99.36&98.99&95.21&90.01&85.97&70.04&53.47&45.12&72.25&56.55&47.73\\
&SA-AET(LI)&100&100&100&98.33&96.63&95.25&66.91&49.22&40.80&69.09&51.92&43.14\\
&SA-AET(LI)+SIA&100&\textbf{100}&99.98&\textbf{99.38}&\textbf{98.78}&\textbf{98.35}&78.58&66.36&59.70&80.41&68.15&60.49\\
&\cellcolor[rgb]{0.925,0.925,0.925}{SADCA}
 &\cellcolor[rgb]{0.925,0.925,0.925}{\textbf{100}}
 &\cellcolor[rgb]{0.925,0.925,0.925}{99.98}
 &\cellcolor[rgb]{0.925,0.925,0.925}{\textbf{99.98}}
 &\cellcolor[rgb]{0.925,0.925,0.925}{97.83}
 &\cellcolor[rgb]{0.925,0.925,0.925}{94.60}
 &\cellcolor[rgb]{0.925,0.925,0.925}{92.65}
 &\cellcolor[rgb]{0.925,0.925,0.925}{\textbf{82.83}}
 &\cellcolor[rgb]{0.925,0.925,0.925}{\textbf{68.89}}
 &\cellcolor[rgb]{0.925,0.925,0.925}{\textbf{61.49}}
 &\cellcolor[rgb]{0.925,0.925,0.925}{\textbf{86.11}}
 &\cellcolor[rgb]{0.925,0.925,0.925}{\textbf{75.30}}
 &\cellcolor[rgb]{0.925,0.925,0.925}{\textbf{68.92}}\\
\midrule
\multirow{10}{*}{TCL}
&SGA&92.77&87.08&83.99&100&99.96&99.96&46.97&28.97&22.15&51.53&33.26&25.55\\
&SGA+SIA&97.22&94.13&92.38&99.98&99.75&99.65&61.76&44.52&36.09&65.21&49.76&41.72\\
&SGA(LI)&85.24&75.72&70.91&100&100&100&38.92&23.27&17.23&45.97&27.49&20.92\\
&SGA(LI)+SIA&99.70&99.28&99.03&100&99.98&99.98&59.12&42.37&35.41&65.49&50.75&42.58\\
&DRA&95.58&91.10&88.66&100&99.94&99.94&57.28&39.90&32.16&62.23&43.55&35.19\\
&SA-AET&98.50&96.60&95.45&100&100&99.96&63.47&46.55&38.64&67.86&49.90&41.52\\
&SA-AET+SIA&97.36&94.48&92.28&99.86&99.38&99.21&72.97&57.65&50.52&75.40&61.77&53.94\\
&SA-AET(LI)&99.20&98.61&98.08&100&100&100&64.37&47.44&40.10&69.30&51.94&44.18\\
&SA-AET(LI)+SIA&\textbf{99.93}&\textbf{99.82}&\textbf{99.66}&100&100&100&81.48&69.75&63.65&84.05&73.85&67.47\\
&\cellcolor[rgb]{0.925,0.925,0.925}{SADCA}
 &\cellcolor[rgb]{0.925,0.925,0.925}{99.56}
 &\cellcolor[rgb]{0.925,0.925,0.925}{98.79}
 &\cellcolor[rgb]{0.925,0.925,0.925}{98.28}
 &\cellcolor[rgb]{0.925,0.925,0.925}{\textbf{100}}
 &\cellcolor[rgb]{0.925,0.925,0.925}{\textbf{100}}
 &\cellcolor[rgb]{0.925,0.925,0.925}{\textbf{100}}
 &\cellcolor[rgb]{0.925,0.925,0.925}{\textbf{83.17}}
 &\cellcolor[rgb]{0.925,0.925,0.925}{\textbf{70.61}}
 &\cellcolor[rgb]{0.925,0.925,0.925}{\textbf{64.17}}
 &\cellcolor[rgb]{0.925,0.925,0.925}{\textbf{88.71}}
 &\cellcolor[rgb]{0.925,0.925,0.925}{\textbf{79.69}}
 &\cellcolor[rgb]{0.925,0.925,0.925}{\textbf{74.38}}\\
\midrule
\multirow{10}{*}{CLIP\textsubscript{ViT}}
&SGA&34.59&18.27&13.99&36.45&19.43&14.11&100&100&100&61.10&43.50&35.83\\
&SGA+SIA&55.22&35.42&28.22&56.48&37.44&29.89&99.77&99.53&99.24&83.12&69.97&62.34\\
&SGA(LI)&31.08&15.34&11.04&33.52&16.85&11.76&100&100&100&56.23&37.63&30.63\\
&SGA(LI)+SIA&60.20&41.82&33.64&62.60&43.64&36.45&100&100&100&89.24&81.00&74.95\\
&DRA&42.84&25.47&19.29&44.60&25.93&19.60&100&100&99.93&69.50&54.46&46.15\\
&SA-AET&50.44&32.98&26.14&51.10&33.16&25.87&100&99.93&99.93&74.10&60.24&52.72\\
&SA-AET+SIA&65.13&47.27&40.80&66.33&49.14&42.06&99.87&99.32&98.95&85.69&74.82&68.74\\
&SA-AET(LI)&57.16&39.60&32.57&57.52&39.47&31.90&100&100&100&80.17&66.50&59.59\\
&SA-AET(LI)+SIA&82.74&70.14&63.89&82.57&70.73&64.38&99.97&99.84&99.83&95.23&90.83&87.51\\
&\cellcolor[rgb]{0.925,0.925,0.925}{SADCA}
 &\cellcolor[rgb]{0.925,0.925,0.925}{\textbf{89.20}}
 &\cellcolor[rgb]{0.925,0.925,0.925}{\textbf{75.84}}
 &\cellcolor[rgb]{0.925,0.925,0.925}{\textbf{68.90}}
 &\cellcolor[rgb]{0.925,0.925,0.925}{\textbf{87.98}}
 &\cellcolor[rgb]{0.925,0.925,0.925}{\textbf{73.91}}
 &\cellcolor[rgb]{0.925,0.925,0.925}{\textbf{67.00}}
 &\cellcolor[rgb]{0.925,0.925,0.925}{\textbf{100}}
 &\cellcolor[rgb]{0.925,0.925,0.925}{\textbf{99.98}}
 &\cellcolor[rgb]{0.925,0.925,0.925}{\textbf{99.98}}
 &\cellcolor[rgb]{0.925,0.925,0.925}{\textbf{97.46}}
 &\cellcolor[rgb]{0.925,0.925,0.925}{\textbf{93.98}}
 &\cellcolor[rgb]{0.925,0.925,0.925}{\textbf{91.48}}\\
\midrule
\multirow{10}{*}{CLIP\textsubscript{CNN}}
&SGA&28.06&15.01&10.68&33.07&16.66&12.02&51.45&32.19&25.27&99.90&99.73&99.64\\
&SGA+SIA&33.63&17.49&12.66&37.57&20.67&15.21&54.41&36.72&38.96&99.73&98.93&97.90\\
&SGA(LI)&28.02&14.03&9.97&31.36&15.77&11.19&48.90&29.57&22.57&99.93&99.90&99.80\\
&SGA(LI)+SIA&32.72&17.19&12.52&37.21&20.55&15.11&56.70&37.86&30.92&100&99.95&99.86\\
&DRA&34.59&18.56&13.93&37.88&21.15&15.98&59.12&99.64&32.71&99.90&40.32&99.41\\
&SA-AET&38.28&21.47&16.60&41.81&24.62&18.03&64.21&44.34&36.46&99.97&99.71&99.50\\
&SA-AET+SIA&41.53&24.65&19.07&46.62&28.24&21.77&64.11&46.90&38.70&99.18&97.96&97.18\\
&SA-AET(LI)&44.74&27.42&21.71&47.45&29.48&23.09&68.52&52.32&44.29&100&100&100.00\\
&SA-AET(LI)+SIA&51.80&33.65&26.95&56.33&38.00&31.07&74.68&59.87&52.33&100&99.93&99.89\\
&\cellcolor[rgb]{0.925,0.925,0.925}{SADCA}
 &\cellcolor[rgb]{0.925,0.925,0.925}{\textbf{60.55}}
 &\cellcolor[rgb]{0.925,0.925,0.925}{\textbf{38.31}}
 &\cellcolor[rgb]{0.925,0.925,0.925}{\textbf{30.89}}
 &\cellcolor[rgb]{0.925,0.925,0.925}{\textbf{63.19}}
 &\cellcolor[rgb]{0.925,0.925,0.925}{\textbf{41.29}}
 &\cellcolor[rgb]{0.925,0.925,0.925}{\textbf{33.10}}
 &\cellcolor[rgb]{0.925,0.925,0.925}{\textbf{79.57}}
 &\cellcolor[rgb]{0.925,0.925,0.925}{\textbf{64.07}}
 &\cellcolor[rgb]{0.925,0.925,0.925}{\textbf{56.17}}
 &\cellcolor[rgb]{0.925,0.925,0.925}{\textbf{100}}
 &\cellcolor[rgb]{0.925,0.925,0.925}{\textbf{100}}
 &\cellcolor[rgb]{0.925,0.925,0.925}{\textbf{100}}\\
\midrule
\end{tabular}
}
\label{IR1510}
\end{table*}

\section{Visualization}
\begin{figure*}[t!]
\centering
\includegraphics[width=0.99\textwidth]{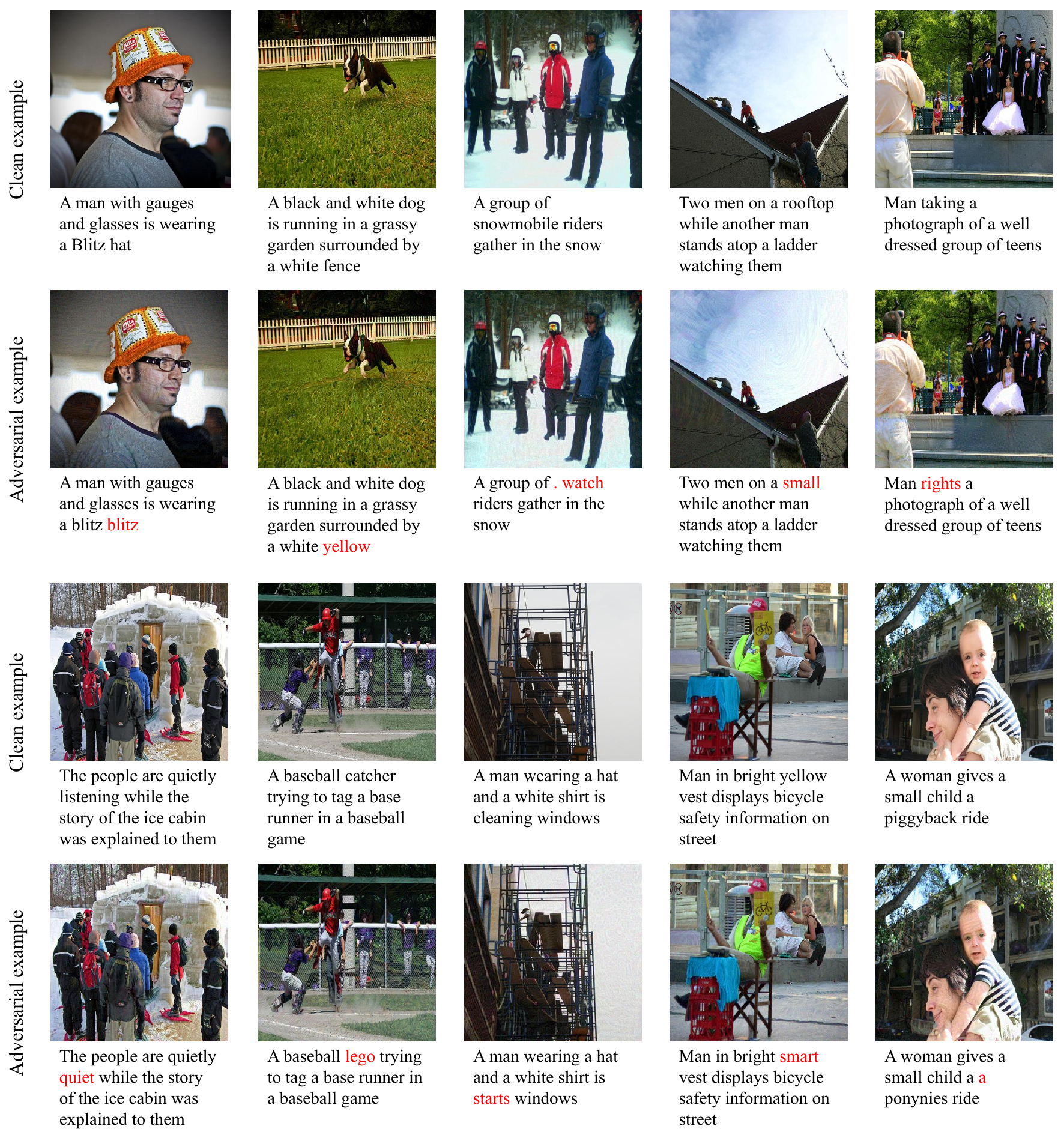}
\caption{Visualization of original images and the corresponding adversarial examples generated by our proposed SADCA.}
\label{AE}
\end{figure*}

\begin{figure*}[t!]
\centering
\includegraphics[width=0.99\textwidth]{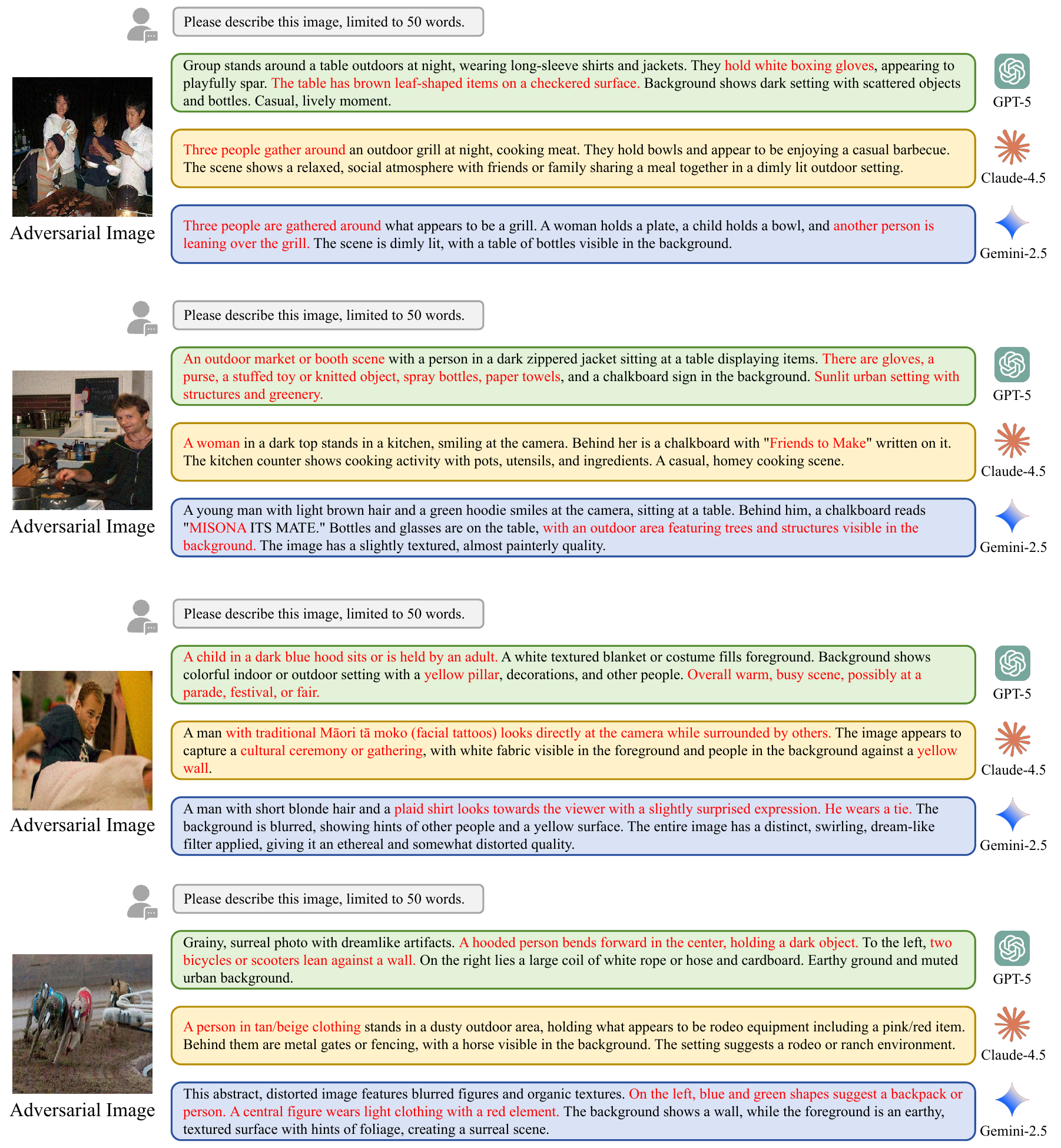}
\caption{Visualization of adversarial images in attacking commercial LVLMs.}
\label{AE_LVLM}
\end{figure*}

Figure \ref{AE} shows the visualization of randomly selected clean examples and adversarial examples.
Figure \ref{AE_LVLM} shows the description of commercial LVLMs for the adversarial images generated by SADCA. 
It can be seen that it is capable of effectively attacking various commercial LVLMs.

\end{document}